\documentclass[manuscript,acmtog,review=false,nonacm=true]{acmart}
\AtBeginDocument{%
  }

\DeclareMathOperator*{\argmin}{arg\,min}

\newcommand{\norm}[1]{\left\lVert#1\right\rVert}

\setcopyright{none}

\begin{document}
\title{Fast Local Neural Regression for Low-Cost, Path Traced Lambertian Global Illumination}

\author{Arturo Salmi}
\orcid{0009-0007-3745-4574}
\affiliation{%
  \institution{Imagination Technologies}
  \country{UK}
}
\email{arturo.salmi@imgtec.com}

\author{Szabolcs Cs\'efalvay}
\orcid{0009-0001-1790-6680}
\affiliation{%
    \institution{Imagination Technologies}
    \country{UK}
}
\email{szabolcs.csefalvay@imgtec.com}

\author{James Imber}
\orcid{0009-0005-1757-9447}
\affiliation{%
    \institution{Imagination Technologies}
    \country{UK}
}
\email{james.imber@imgtec.com}


\renewcommand{\shortauthors}{Salmi et al.}

\begin{teaserfigure}
 \includegraphics[width=\textwidth]{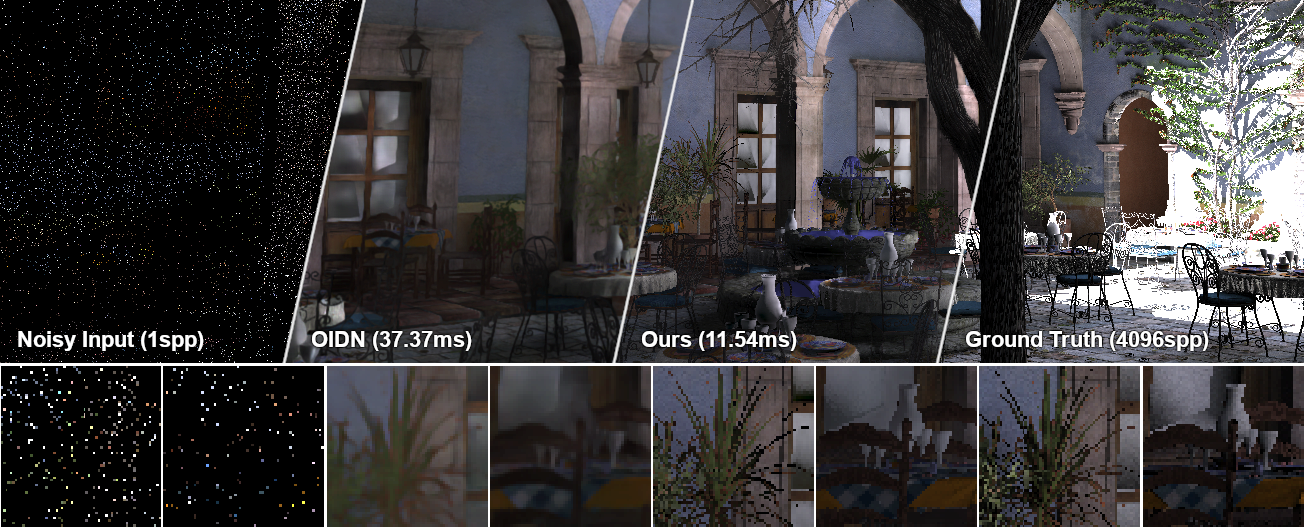}
 \centering
  \caption{Our proposed Fast Local Neural Regression (FLNR) spatial denoising algorithm achieves convincing global illumination for Lambertian scenes at very low sample counts (1 indirect sample per pixel). By using a neural network to generate an enhanced guide image for a novel windowed linear regression method, we improve clarity and robustness over previous deep learning based methods. Execution times measured on an Nvidia RTX 2080 Ti.}
\label{fig:teaser}
\end{teaserfigure}

\begin{abstract}
  Despite recent advances in hardware acceleration of ray tracing, real-time ray budgets remain stubbornly limited at a handful of samples per pixel (spp) on commodity hardware, placing the onus on denoising algorithms to achieve high visual quality for path traced global illumination.
  Neural network-based solutions give excellent result quality at the cost of increased execution time relative to hand-engineered methods, making them less suitable for deployment on resource-constrained systems.
  We therefore propose incorporating a neural network into a computationally-efficient local linear model-based denoiser, and demonstrate faithful single-frame reconstruction of global illumination for Lambertian scenes at very low sample counts (1spp) and for low computational cost.
  Other contributions include improving the quality and performance of local linear model-based denoising through a simplified mathematical treatment, and demonstration of the surprising usefulness of ambient occlusion as a guide channel.
  We also show how our technique is straightforwardly extensible to joint denoising and upsampling of path traced renders with reference to low-cost, rasterized guide channels.
\end{abstract}

\maketitle

\section{Introduction}
With the commoditisation of ray tracing acceleration hardware in commercial GPUs, interactive- and real-time applications of 
Monte Carlo ray tracing methods are increasing in popularity, including physics-based light transport algorithms such as 
path tracing~\cite{Kajiya1986}. Due to slow convergence with increasing sample count, path tracing for global illumination 
often requires thousands of samples per pixel (spp) to render an acceptably low-noise frame, which despite 
the availability of dedicated ray tracing hardware is still prohibitively expensive for real-time applications. 
Image denoising methods are therefore commonly employed to achieve interactive frame rates at sufficient quality.

Concurrently with wider adoption of ray tracing, the greater availability of AI acceleration hardware and recent advancements in neural image processing techniques have had a considerable impact on Computer Graphics. 
For example, techniques such as neural frame upsampling~\cite{Liu2020} \cite{Chowdhury2022}, frame interpolation~\cite{Briedis2023}, and neural shading~\cite{Zeltner2024} \cite{muller2021} hold the promise 
of drastically reducing the size of graphics workloads, simultaneous with increasing the realism~\cite{Richter2021} of outputs.
Path tracing similarly stands to gain from these new capabilities, particularly from recent denoising methods which use
neural networks~\cite{Chaitanya2017} \cite{Attila2024}, potentially combined with upscaling~\cite{Thomas2022} \cite{Zhang2023}, to give excellent quality at low spp budgets. Such neural methods 
significantly reduce the costs associated with using dedicated ray tracing hardware while better preserving relevant visual details such as global illumination and soft shadows.

Although some path traced image denoisers can achieve real-time performance on commodity hardware, there is still scope for further improvements to reduce latency and
improve image quality, particularly at the lowest sample counts.
Neural networks are often less computationally efficient than a carefully constructed hand-engineered algorithm exploiting domain knowledge~\cite{Salmi2023}. 
We show that a robust, high-quality hand-engineered spatial denoiser, when augmented with a neural network, can achieve state-of-the-art denoising results at
a fraction of the execution time of more conventional neural network-based regression models (Figure~\ref{fig:teaser}).
In summary, we present:
\begin{itemize}
  \item A simple, numerically stable and differentiable windowed linear regression algorithm (Fast Local Regression) capable of faithful reconstruction of 1080p path traced images of Lambertian scenes at very low sample counts (e.g. 1spp), at sub-millisecond execution time on commodity desktop hardware.
  \item Further improvements of the quality of this model by inclusion of a neural network (Fast Local Neural Regression), with faster execution on desktop class GPUs, and at higher quality, than comparable neural network based methods.
\end{itemize}

\begin{figure*}[h!]
  \centering
  \includegraphics[width=1\linewidth]{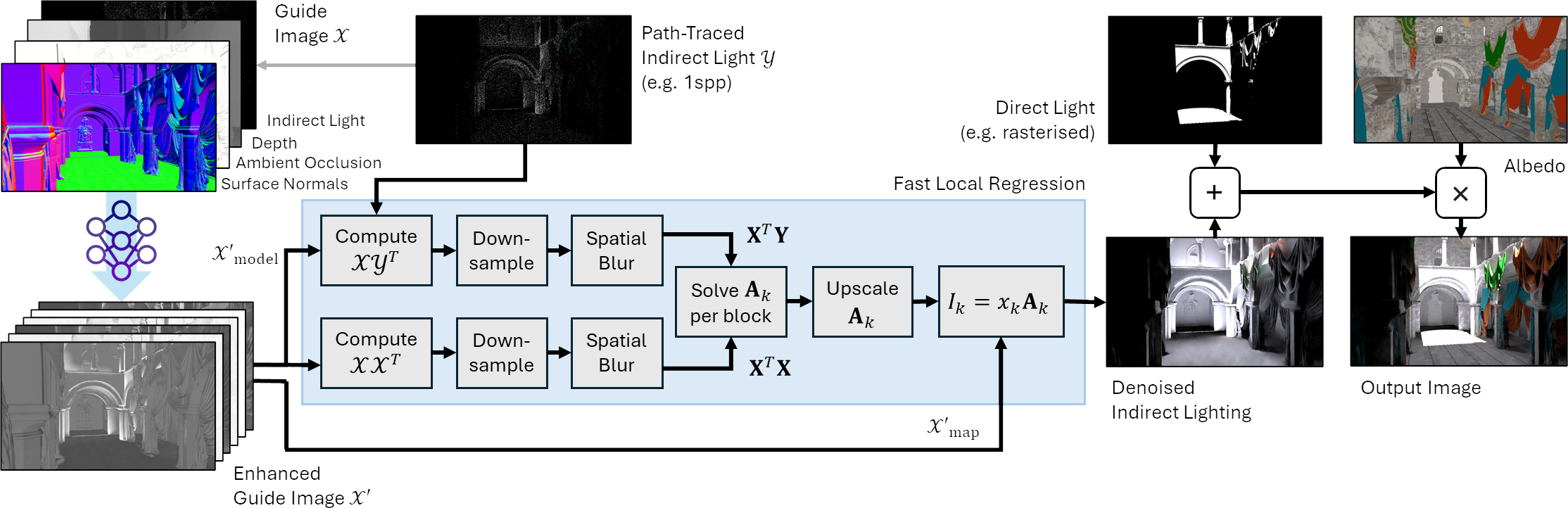}
  \caption{
    In Fast Local Neural Regression, the guide image and indirect light are passed into a neural network, which extracts structural information for the linear model to exploit. Our Fast Local Regression algorithm efficiently fits noise-free guide channels (e.g. rasterised at low cost) to a noisy, path traced input image. Optimisations include fitting the model $\textbf{A}_k$ to 8x8 blocks, and use of cheap downsampling and separable blurs to collect local moments.
  }
  \label{fig:pipeline}
\end{figure*}

\section{Related Work}

Although methods to reduce sample variance in Monte Carlo rendering have been explored for decades~\cite{Veach1998} \cite{Zwicker2015},
interest in real-time denoising solutions has grown with the popularity of ray tracing in commercial GPUs, in turn driven by the 
availability of dedicated ray tracing hardware.  

Initial interactive frame rate solutions used edge-avoiding à-trous wavelets~\cite{Dammertz2010}, which are multi-resolution edge-stopping 
functions based on auxiliary images (surface normals, depth etc.). Later, Schied et al. extend this wavelet-based approach 
by adding temporal reprojection to accumulate more samples over time~\cite{Schied2017}, significantly improving visual quality and temporal 
stability. Temporal accumulation reduces variance considerably, but overreliance also introduces temporal lag, making the results less 
reactive to changes in lighting and disocclusions. Later work addresses this by using temporal gradients to reset or down-weight temporal history 
where necessary~\cite{Schied2018}.

There is considerable precedent in the literature for separating aspects of appearance into distinct signals for separate processing.
Zeng et al.~\cite{Zeng2021} treat shading, shadows, glossy reflections and diffuse indirect illumination separately to 
accumulate samples more accurately. The interactive denoiser used in Unreal Engine 4 is a further example, which 
separates the signal components and, with reference to geometry information, constructs a different anisotropic filter for each 
component~\cite{Liu2019}. Similarly to our approach, Bauszat et al. adopt a local linear model which fits surface normals and depth 
to the noisy indirect illumination estimate~\cite{Bauszat2011} for global illumination.

Neural network-based solutions typically offer higher quality than handcrafted methods but are less computationally efficient \cite{Salmi2023}. 
To address these shortcomings, recent work has been focused on using small neural networks to augment handcrafted solutions. 
Kalantari et al.~\cite{Kalantari2015} train a multi-layer perceptron to output the optimal parameters of cross-bilateral and non-local means filters. 
Bako et al.~\cite{Bako2017} use a CNN to construct per-pixel linear kernels to denoise diffuse and specular components. Although these solutions 
produce impressive results, they are still too expensive for real-time applications. 

As a cheaper alternative, U-Net-based denoising autoencoders achieve a large receptive field at relatively low cost. Chaitanaya et al.~\cite{Chaitanya2017}
propose a recurrent U-Net denoiser which directly outputs the denoised frame and can be executed at interactive frame rates on an 
Nvidia Titan X. Hasselgren et al.~\cite{Hasselgren2020} use a recurrent U-Net to generate a feature map that is used for spatiotemporal adaptive sampling 
and to construct per-pixel linear denoising kernels. To reduce the cost associated with constructing per-pixel kernels, Işik et al.~\cite{Isik2021}
generate per-pixel features that are used to construct adaptive dilated kernels. Balint et al.~\cite{Balint2023} use a series of lightweight U-Nets 
to construct a low-pass pyramid filter, where the kernel constructor is also trained to separate the input radiance between pyramidal layers
as an alternative to classical downsampling and upsampling approaches, which are often prone to aliasing. In common with the aforementioned
methods, our approach uses a U-Net, although it in our case it is integrated into a local linear model for improved computational efficiency, rather
than generating kernels or output pixels directly. Thomas et al.~\cite{Thomas2022} propose a temporal approach using a U-Net to output an upscaled frame,
with intermediate features of the network used to estimate per-pixel linear kernels for application to the noisy radiance image.

Guided filtering~\cite{He2010} introduced local linear models (more properly "local affine models") to the image processing
literature. This quickly found application to denoising of indirect lighting in the aforementioned work of Bauszat et al.~\cite{Bauszat2011}. Important
performance optimisations, particularly fitting of the model at low resolution, were proposed by He and Sun~\cite{He2015}.
Our Fast Linear Regression starts from the same intuition, namely that an input signal can be reconstructed as a least-squares optimal affine
combination of multiple guide signals, but by starting from first principles, we identify a simpler, faster, more robust and higher-quality
way to achieve this.
\begin{itemize}
  \item We downsample the statistical moment tensor after collecting outer products, rather than the input images as in the fast guided filter.
This is important for maintaining result quality, since downsampling the input is not information-preserving.
  \item FLR is comparatively simple and easy to understand, which lends itself to straightforward differentiable implementation in
PyTorch (Figure \ref{fig:pytorch_code}). FLR with multiple guide and target channels is no more complex than the single-channel case.
  \item The original guided filter requires two filtering and two integral image generation steps.
We combine this into a single step of blurring the moments, leading to a performance improvement.We show that a single blur pass only is required,
and that there is benefit to weighting the contribution of centre pixels in the local region more than distant pixels.
\end{itemize}

\section{Method}

Light transport is defined by the non-emissive part of the rendering equation~\cite{Kajiya1986}:

\begin{equation}
  L_o(\omega_o) = \int_{\Omega} L_{i} \left(x, \omega_{i}\right)f_{s}\left(x, \omega_{i}, \omega_{o}\right)\left(\omega_{i} \cdot n_{x}\right) d\omega_{i}
  \label{equ:rendering}
\end{equation}

where $\omega_{o}$ is the direction of the ray reflected at intersected point $x$, $\omega_{i}$ is the direction of the incoming ray, $L_{i}$ is the incoming radiance, $f_{s}$ is the bidirectional scattering distribution function (BSDF) and $n_{x}$ represents the normal at position $x$.
Path tracing evaluates this otherwise-intractable integral using Monte Carlo sampling. The denoising problem arises when insufficient samples are available, resulting in unconverged, high-noise images. For Lambertian scenes, this simplifies to the following:

\begin{equation}
  L_o = f_s(x) \int_{\Omega} L_{i} \left(x, \omega_{i}\right)\left(\omega_{i} \cdot n_{x}\right) d\omega_{i}
\end{equation}

Furthermore, in common with~\cite{Bauszat2011}, the linearity (superposition) of this equation can be exploited to split into direct and indirect lighting, with the majority of noise concentrated in the latter:

\begin{equation}
\begin{split}
  L_o = f_s(x) \bigg( \int_{\Omega} L_\text{direct} \left(x, \omega_{i}\right)\left(\omega_{i} \cdot n_{x}\right) d\omega_{i} +\\ \int_{\Omega} L_\text{indirect} \left(x, \omega_{i}\right)\left(\omega_{i} \cdot n_{x}\right) d\omega_{i} \bigg)
\end{split}
\end{equation}

Here, $L_\text{direct}$ integrates over the direct radiance reflected from primary ray intersections, while $L_\text{indirect}$ integrates over the indirect radiance reflected from secondary ray intersections.
We use a noise-free estimate of the direct radiance and restrict the task of our method to denoising of $L_\text{indirect}$ for indirect diffuse radiance in Lambertian scenes.

We describe our method with reference to Figure~\ref{fig:pipeline}.
The core of our method (blue box) is Fast Local Regression (FLR), described in detail in Section~\ref{sec:fast_local_regression},
with mathematical background covered in Section~\ref{sec:fresh_look} and the Appendix.
This is a capable denoising algorithm in its own right, taking inexpensively rendered (e.g. rasterised), noise-free guide channels as an
auxiliary input. The guide channels are then fitted to the noisy path-traced image using a local weighted least
squares algorithm, resulting in the denoised indirect lighting. This is then recombined with the direct light and albedo,
yielding the final output image.

The main shortcoming of a regression based filter is that it can only adapt to structure present in the guides.
It effectively misidentifies changes in illumination that do not covary with the guides as noise,
removing cast shadows within the receptive field.
To overcome this limitation we transform guide channels with a neural network in our Fast Local Neural Regression~(FLNR) algorithm,
as shown in Figure~\ref{fig:pipeline} and described in Section~\ref{sec:fast_local_neural_regression}. This builds on a differentiable implementation of FLR, allowing gradients
to be backpropagated into the neural network to enable training. The neural network learns to output enhanced guide channels,
containing extracted structural information to facilitate denoising of challenging features, such as cast shadows. Feeding input
noisy lighting in as an additional guide allows the network to adapt the enhanced guides to maximise quality for given lighting conditions.
By limiting the complexity of the task the neural network has to perform
we can reduce its complexity and execution time without creating artefacts,
resulting in a robust, content-adaptive denoiser for diffuse light.

\begin{figure}[t]
  \centering
  \hrule height 1pt depth 0pt width \linewidth \relax
  \begin{footnotesize} 
\begin{verbatim}
def local_regression(X, Y, eps=0.000001, kernel_size=41, std=10.0):
  XX = torch.einsum("ikl,jkl->ijkl", X, X)
  XY = torch.einsum("ikl,jkl->ijkl", X, Y)
  t = torchvision.transforms.GaussianBlur(kernel_size, std)
  XX = t(XX).permute([2, 3, 0, 1])
  XY = t(XY).permute([2, 3, 0, 1])
  XXXY = torch.linalg.solve(XX + eps * torch.eye(X.shape[0]), XY)
  return torch.einsum("jkl,klji->ikl", X, XXXY)
\end{verbatim}
  \end{footnotesize}
  \hrule height 1pt depth 0pt width \linewidth \relax
  \caption{
    A basic, differentiable implementation of the core local linear algorithm in PyTorch, with basic regularisation.
    It returns a $\left(3 \times H \times W\right)$ denoised RGB image.
  }
  \label{fig:pytorch_code}
\end{figure}

\subsection{A Fresh Look at Local Linear Models}
\label{sec:fresh_look}

In this section we introduce a straightforward mathematical formulation of local linear models, which is intuitively easier than
previous descriptions of related methods, and lends itself to differentiable implementation in deep learning frameworks such as PyTorch.

We intend to reconstruct indirect lighting $\mathcal{I} \in \mathbb R^{\left(H \times W \times 3\right)}$
from a noisy RGB indirect radiance signal tensor $\mathcal{Y} \in \mathbb R^{\left(H \times W \times 3\right)}$
with reference to a guide tensor $\mathcal{X} \in \mathbb R^{\left(H \times W \times Q\right)}$,
where $H$ and $W$ represent the height and width of the input image respectively, and $Q$ is the number of guides used.
For each tensor we define a 
square window $T_{\textit{k}}$ centered around pixel $\textit{k}$. The values of $\mathcal{Y}$ and $\mathcal{X}$
contained in $T_{\textit{k}}$ are used to define respective signal $\textbf{Y}_{\textit{k}} \in \mathbb R^{\left(N \times 3\right)}$ 
and guide $\textbf{X}_{\textit{k}} \in \mathbb R^{\left( N \times \left( Q + 1\right)\right)}$ matrices (Equations~\ref{equ:matX}~and~\ref{equ:matY}). 
The first column (guide) in $\textbf{X}_{\textit{k}}$ always consists of 1s, that is, $\textbf{X}_{i,0}=1$, to facilitate modelling of constant bias (offset).
Each row of these matrices corresponds to a pixel location in $T_{\textit{k}}$, while each column represents a different 
channel of the respective tensors.

\begin{equation}
  \textbf{X}_{\textit{k}} =  
  \begin{bmatrix}
   \;\; 1   & \textbf{X}_{0,1}   & \textbf{X}_{0,2}   & \dots  & \textbf{X}_{0,Q}   \\
   \;\; 1   & \textbf{X}_{1,1}   & \textbf{X}_{1,2}   & \dots  & \textbf{X}_{1,Q}   \\
   \;\; 1   & \textbf{X}_{2,1}   & \textbf{X}_{2,2}   & \dots  & \textbf{X}_{2,Q}   \\
   \;\; \vdots             & \vdots             & \vdots             & \ddots & \vdots               \\
  \;\; 1 & \textbf{X}_{N-1,1} & \textbf{X}_{N-1,2} & \dots  & \textbf{X}_{N-1,Q} 
  \end{bmatrix} 
  \label{equ:matX}
\end{equation}

\begin{equation}
  \textbf{Y}_{\textit{k}} =  
  \begin{bmatrix}
    \textbf{R}_{0}   & \textbf{G}_{0}   & \textbf{B}_{0}   \\
    \textbf{R}_{1}   & \textbf{G}_{1}   & \textbf{B}_{1}   \\
    \textbf{R}_{2}   & \textbf{G}_{2}   & \textbf{B}_{2}   \\
    \vdots           & \vdots           & \vdots           \\
    \textbf{R}_{N-1} & \textbf{G}_{N-1} & \textbf{B}_{N-1} 
  \end{bmatrix} 
  \label{equ:matY}
\end{equation}

For each $T_{\textit{k}}$, the objective is to determine the parameter matrix 
$\textbf{A}_{\textit{k}} \in \mathbb R^{\left(\left(Q+1\right) \times 3\right)}$ that maps $\textbf{X}_{\textit{k}}$ to 
$\textbf{Y}_{\textit{k}}$ as closely as possible in a least-squares sense, i.e. such that $\textbf{Y}_{\textit{k}} \approx \textbf{X}_{\textit{k}}\textbf{A}_{\textit{k}}$.
By determining $\textbf{A}_{\textit{k}}$, we can express the noisy radiance signal $\textbf{Y}_\textit{k}$ as a linear 
combination of the noise-free guides $\textbf{X}_{\textit{k}}$, which yields the denoised value for pixel $\textit{k}$.

\begin{equation}
  \textbf{A}^{*}_{\textit{k}} = \argmin_{\textbf{A}_{\textit{k}}}\norm{
    \textbf{X}_{\textit{k}}\textbf{A}_{\textit{k}} - \textbf{Y}_{\textit{k}}
  }^{2}_{2}
\end{equation}

$\textbf{A}^{*}_{\textit{k}}$ has the very simple closed form solution:

\begin{equation}
  \textbf{A}^{*}_{\textit{k}} = \left(\textbf{X}^{T}_{\textit{k}}\textbf{X}_{\textit{k}}\right)^{-1} 
  \textbf{X}^{T}_{\textit{k}}\textbf{Y}_{\textit{k}}
\label{equ:closed_form}
\end{equation}

The solution $\textbf{A}^{*}_{\textit{k}}$ can then be used to determine the denoised pixel value $\textbf{I}_{\textit{k}}$ as:

\begin{equation}
  \textit{\textbf{I}}_{\textit{k}} = \textbf{x}_{k}\textbf{A}^{*}_{\textit{k}}
\end{equation}

Where $\textbf{x}_{k} \in \mathbb R^{Q+1}$ is the row of the guidance matrix $\textbf{X}_{\textit{k}}$ corresponding to the pixel $\textit{k}$.
As $\textbf{A}_{\textit{k}}$ only applies for a given window $T_{\textit{k}}$, it is necessary to find $\textbf{A}_{\textit{k}}$ for every
pixel location $\textit{k}$ in the input tensors $\mathcal{X}$ and $\mathcal{Y}$.

For modest numbers of guide channels (e.g. 6), the symmetric matrix inverse is not a performance bottleneck.
Together with downscaling of moments (Section~\ref{sec:fast_local_regression}), our matrix inversion completes
in approximately 18 microseconds on an Nvidia RTX 2080 Ti for a 1080p frame.
In practical applications, it is generally necessary to regularise the problem to avoid numerical stability issues (and possible
rank defficiency) in Equation~\ref{equ:closed_form}. Although simple Tikhonov regularisation does work, our full regularisation
solution for best results is detailed in the Appendix.

\subsection{Fast Local Regression}
\label{sec:fast_local_regression}

As with He and Sun~\cite{He2015}, we exploit redundant computation between overlapping windows. Given outer product tensors\\
$\mathcal{X}\mathcal{X}^T \in \mathbb R^{\left(H \times W \times \left(Q+1\right) \times \left(Q+1\right)\right)}$ and
$\mathcal{X}\mathcal{Y}^T \in \mathbb R^{\left(H \times W \times \left(Q+1\right) \times 3\right)}$, blurring with
an efficient $O(1)$ box filter yields exactly the per-pixel matrices $\textbf{X}^T\textbf{X}$ and $\textbf{X}^T\textbf{Y}$
in Equation~\ref{equ:closed_form}.
Unlike the guided filter however, we discern no advantage to blurring the model.

Futher quality improvements are achieved with blur kernels that weight pixels near the centre preferentially to pixels at the edge of
the window. Replacing the box filter wih a separable Gaussian window is a cheap option ($O(K)$).
This is similar in approach to \cite{cleveland1979robust}, and corresponds to solving the weighted least squares problem:

\begin{equation}
  \textbf{A}^{*}_{\textit{k}} = \argmin_{\textbf{A}_{\textit{k}}}\norm{
    \text{diag}(\textbf{w})^{\frac{1}{2}}\left(\textbf{X}_{\textit{k}}\textbf{A}_{\textit{k}} - \textbf{Y}_{\textit{k}}\right)
  }^{2}_{2}
\end{equation}

Where the vector $\textbf{w}$ is the per-pixel weighting within the local window. In a differentiable implementation
(as in in Figure~\ref{fig:pytorch_code}), the blur radius \(\sigma\) as well as any regularisation parameters are
learned with backpropagation.

Although the above method is sufficient to compute a denoised image, it is still inefficient, since
it requires calculation of $\textbf{X}_{k}^{T}\textbf{X}_{k}$, its inverse, and $\textbf{X}_{k}^{T}\textbf{Y}_{k}$ at every pixel position with windowing.
Following the fast guided filter~\cite{He2015}, we exploit the large standard deviation of the blur kernel \(\sigma\), and decompose the blur into a downsample
(in our case, 8x), followed by a smaller blur (in our case with standard deviation \(\frac{\sigma}{8}\)).
However, we downsample the moment tensor rather than the input image, which is important for maximising image quality.
This means that we produce a linear model for every 8x8 pixel block, and then bilinearly interpolate model parameters.
This allows us to reduce the cost of blurring and matrix inversion by a factor of 64 with no discernable loss of quality.

\begin{figure}[t]
  \centering
  \includegraphics[width=1\linewidth]{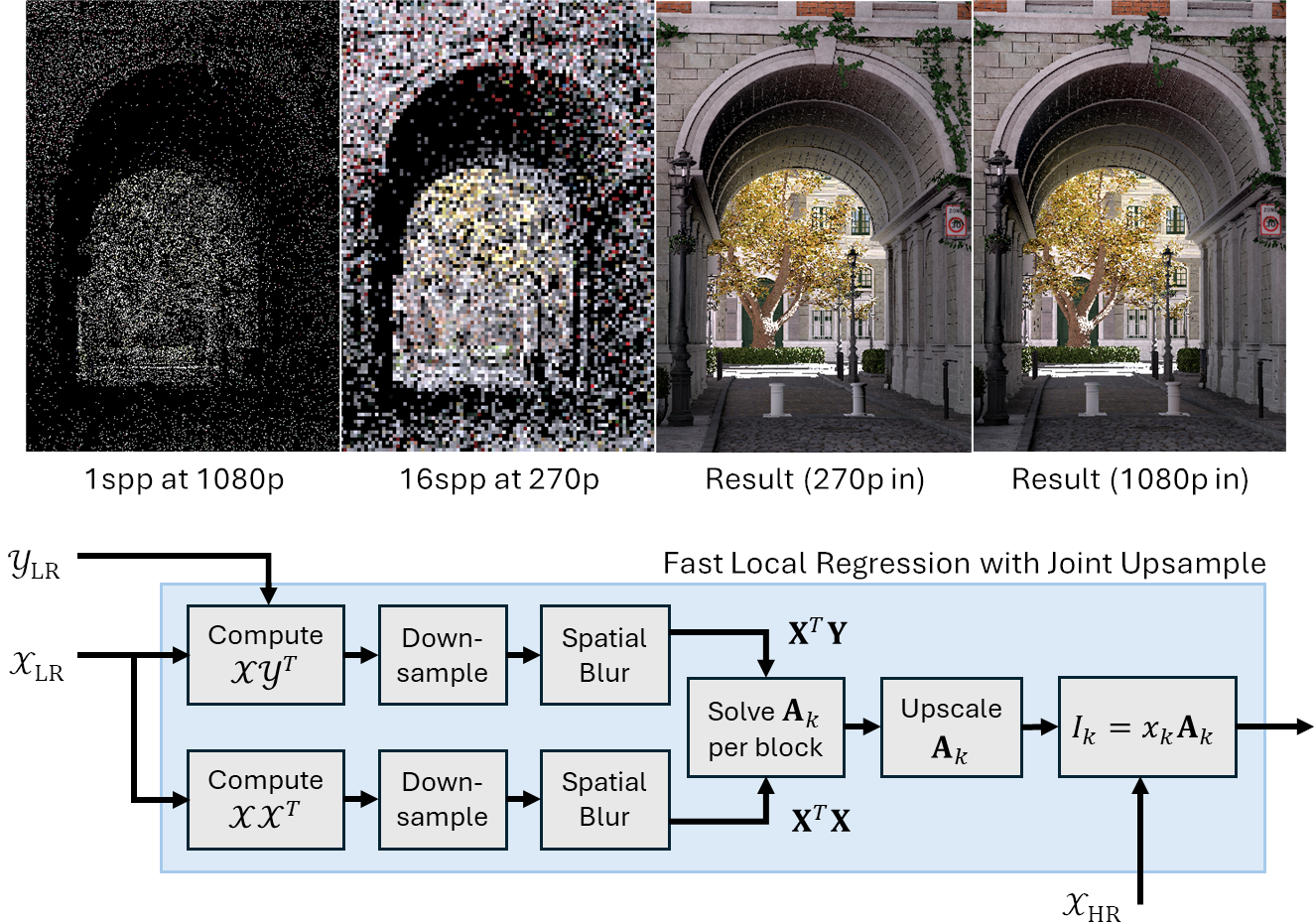}
  \caption{
    Cheaply rasterising two sets of corresponding guide channels, one at low resolution and one at high resolution, allows indirect lighting to be
path traced at low resolution and jointly denoised and upsampled with minimal loss in quality.
  }
  \label{fig:joint_upsample}
\end{figure}

The method is implemented in four optimised OpenCL compute kernels:
\begin{enumerate}
\item Outer products and downsample (strided box filter).
\item Separable Gaussian blur (applied separately for $\textbf{X}_{k}^{T}\textbf{X}_{k}$ and $\textbf{X}_{k}^{T}\textbf{Y}_{k}$).
\item Solver: evaluation of $\left(\textbf{X}_{k}^{T}\textbf{X}_{k}\right)^{-1}\textbf{X}_{k}^{T}\textbf{Y}_{k}$.
\item Upsample and model application.
\end{enumerate}
These stages are split to minimise intermediate bandwidth consumption, by ensuring that all intermediate results are at low resolution.

\subsubsection{Joint Denoising and Upsampling}

Since we reduce the resolution at which we produce the moment matrices \(\textbf{X}_{k}^{T}\textbf{X}_{k}\),
as a further optional optimisation it is also possible to reduce the resolution of the source image as well.
That is, instead of path tracing a full resolution image and producing moment matrices \(\textbf{X}_{k}^{T}\textbf{X}_{k}\)
at an 8x reduced resolution, we can for example path trace a 4x lower resolution source image
followed by a 2x downscaling of the moment tensors with minimal loss of captured information.
This allows the resolution of the input to be traded against the ray budget per pixel.
It can be advantageous to increase the number of bounces per pixel in some scenes, particularly
where high-order backscatter is a significant contributor to lighting.
Furthermore, in some rendering flows, such an approach may lead to substantial savings in bandwidth, power and execution time.
An example crop is shown in Figure~\ref{fig:joint_upsample}, with an overview of the modified FLR method.

\subsubsection{Choosing Guides}

\begin{figure}[t]
  \centering
  \includegraphics[width=1\linewidth]{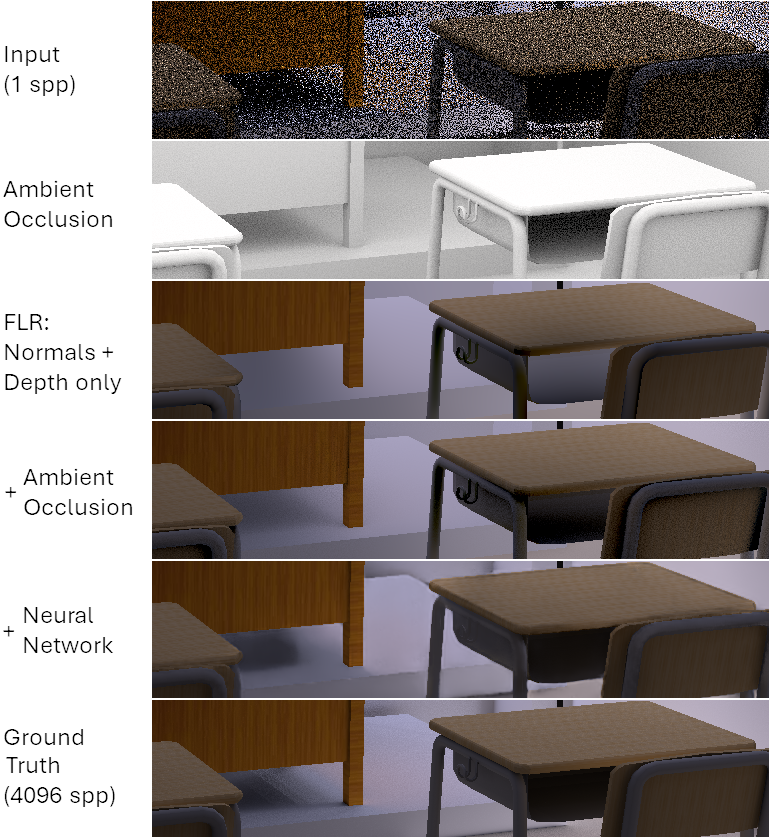}
  \caption{
    Effects of introducing aspects of the proposed method. FLR with surface normal and depth guides correctly reconstructs attached shadows, but typically misses cast shadows. These are recovered with the introduction of ambient occlusion as a guide channel, and neural network generated enhanced guides. The neural network also consumes the noisy input.
  }
  \label{fig:ao}
\end{figure}

In common with previous literature, we confirm the usefulness of depth and surface normals as
useful auxiliary inputs. Although with structural guides alone the guided filter can reconstruct a majority of the scene details, they do 
not present information relating to the likely locations of cast shadows in manner accessible to the model, which 
often results in FLR failing to restore global illumination details such as indirect cast
shadows. Using a precomputed ambient occlusion image proves to be a useful additional 
guide channel, since it contains information about the likely locations and shapes of indirect cast shadows.
An example of the improvement obtained by including this guide is shown in Figure~\ref{fig:ao}.

\begin{figure*}[h!]
  \centering
  \includegraphics[width=1\linewidth]{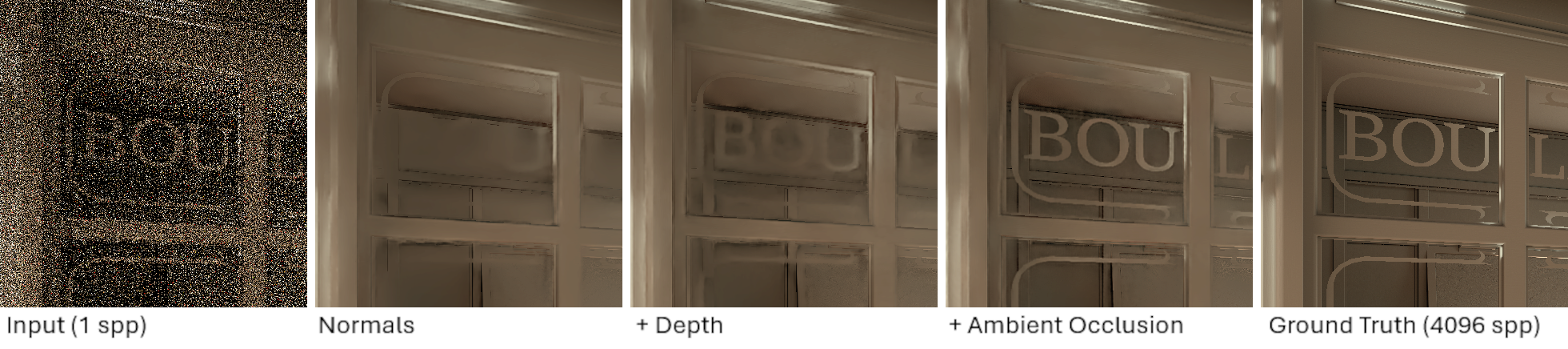}
  \caption{
    Effects of using different combinations of guide channels for the network input in FLNR. All the examples include the noisy input radiance as an additional input.
    Although input radiance and surface normals are sufficient to get a denoised estimate of the frame, the filter looses a great amount of the structural information of the scene.
    Introducing depth and, in particular, ambient occlusion helps the filter restoring the lost structural information, while also improving the quality of denoised shadows.
  }
  \label{fig:FLNR_auxiliary_inputs}
\end{figure*}

\subsection{Fast Local Neural Regression}
\label{sec:fast_local_neural_regression}

The neural network is trained to produce enhanced guide images $\mathcal{X}'$ from the original guide channels $\mathcal{X}$ and the noisy input radiance $\mathcal{Y}$, which are concatenated channel-wise before being fed into the network.
We found it advantageous to output two variants of the enhanced guide channels from the same network, such that
$\mathcal{X}'$ comprises $\mathcal{X}_\text{model}'$ (the set of guide channels we produce the regression model from),
and $\mathcal{X}_\text{map}'$ (the set of guide channels we map to form the output image using the regression model).
In other words, when using a $Q$-dimensional regression model, the network outputs $2Q$ guide channels.
For the input guides fed into the networks, we use the same combination of auxiliary channels adopted for FLR.
An example of the contribution of each guide source for FLNR is shown in Figure~\ref{fig:FLNR_auxiliary_inputs}.

We adopted a U-Net architecture~\cite{Ronneberger2015}, as previous literature showed that this is a good choice for image-to-image 
mapping tasks, including path tracing denoising~\cite{Thomas2022} \cite{Balint2023} \cite{Chaitanya2017} \cite{Hasselgren2020}.
Our layer blocks are designed so that the network can maximise visual quality while also being lightweight enough to be used in real-time applications:

\begin{itemize}
\item A $3x3$ and a $1x1$ convolution projection layer are used as input layers to extract initial features from the network input.
\item Each encoder block uses two $3x3$ convolutional layers followed by a LeakyReLU activation layer. The second convolution strides by 2 to achieve downsampling
(we found separate max pooling for this purpose to be detrimental to result quality and execution efficiency).
\item Decoder blocks receive the features from the previous block, upsampled by a factor of 2. These are concatenated with the corresponding encoder feature map with the same resolution from the encoder, then passed into two $3x3$ convolutional layers followed by a LeakyReLU activation function.
The skip connection is crucial as it helps considerably with detail preservation and recovery in the denoised radiance.
\item Finally, the features obtained from the last decoder block are passed into a $3x3$ and a $1x1$ convolutional layer, and a sigmoid function is used as the output activation layer. 
We found that using the sigmoid function as the output activation significantly improves convergence and stability during training.
\end{itemize}

Our network does not use batch normalisation: although it accelerates convergence, we found it detrimental for the visual quality of the denoised results. 
In future work, we will experiment with layer normalisation instead.

The network is trained in a supervised manner by minimising the SMAPE error between the enhanced filter output and the ground truth, noise-free frame rendered at 4096 spp. 
Training is performed on 512x512 random crops. Flips are used as data augmentation.

To minimise execution time, inference is performed in 16 bit float, and PyTorch offline compilation is used. We expect further gains to be achievable from static graph compilation.

\begin{table*}
  \begin{tabular}{l||ccc|cccc|c}
    \toprule
    \textbf{Method}                  & \textbf{Normals} & \textbf{Geometry} & \textbf{AO} & \textbf{PSNR} $\uparrow$ & \textbf{SSIM} $\uparrow$ & \textbf{FLIP} $\downarrow$ & \textbf{rMSE} $\downarrow$ & \textbf{Runtime (ms)} $\downarrow$\\
    \midrule
    Guided Filter \cite{Bauszat2011} & \checkmark       & \checkmark        &             & 34.925                   & 0.890                    & 0.061                      & 0.024                      & 11.1* \\
    SVGF \cite{Schied2017}           & \checkmark       & \checkmark        &             & 25.924                   & 0.540                    & 0.153                      & 0.070                      & 10.950\\
    BMFR \cite{Koskela2019}          & \checkmark       & \checkmark        &             & 33.194                   & 0.834                    & 0.075                      & 0.028                      & 1.731\\
    NBG \cite{Xiaoxu2020}            & \checkmark       & \checkmark        &             & 33.458                   & 0.828                    & 0.074                      & 0.029                      & 196.614\\
    MKPCN \cite{Xiaoxu2020}          & \checkmark       & \checkmark        &             & 26.933                   & 0.603                    & 0.125                      & 0.057                      & 170.401\\
    OIDN \cite{Attila2024}           & \checkmark       &                   &             & 36.316                   & 0.885                    & 0.055                      & 0.021                      & 37.370\\
    ODDN \cite{Chaitanya2017}        & \checkmark       &                   &             & 34.418                   & 0.813                    & 0.076                      & 0.026                      & 7.313\\
    FLR (Ours) w/o AO                & \checkmark       & \checkmark        &             & 35.334                   & 0.898                    & 0.055                      & 0.023                      & - \\
    FLR (Ours)                       & \checkmark       & \checkmark        & \checkmark  & 35.863                   & 0.897                    & 0.052                      & 0.021                      & \textbf{0.636}\\
    FLNR (Ours)                      & \checkmark       & \checkmark        & \checkmark  & \textbf{37.238}          & \textbf{0.904}           & \textbf{0.046}             & \textbf{0.018}             & 11.542\\
  \bottomrule
\end{tabular}
\caption{
  Comparison of the evaluated methods on 1-spp input frames and their respective required runtime to denoise a 1080p frame.
  "Geometry" means that either depth or world space coordinates are used.
  All methods also use albedo (reflectance), either directly as an auxiliary input (OIDN and ODDN), or to facilitate separate denoising of lighting.
  The denoising quality is evaluated using a series of well-established reference metrics.
  Best scores/runtime are denoted in bold.
  $\uparrow$/$\downarrow$ indicate that higher/lower scores are better.
  * Rescaled for resolution and FLOPs from reported value.
}
\label{tab:results}
\end{table*}

\begin{figure*}[h!]
  \begin{tabular}{ll}
  \includegraphics[scale=0.49]{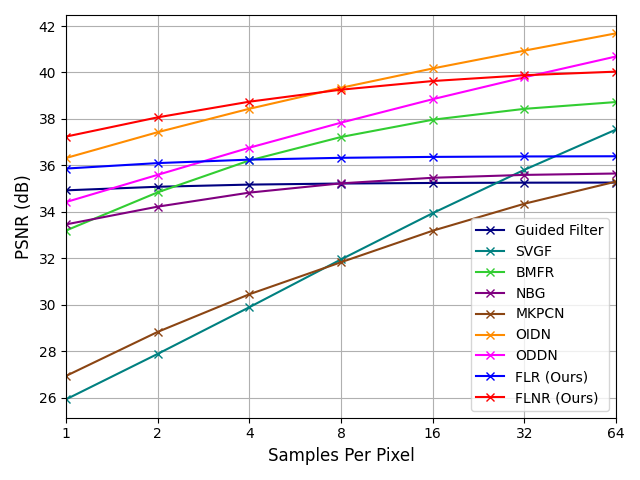}
  &
  \includegraphics[scale=0.49]{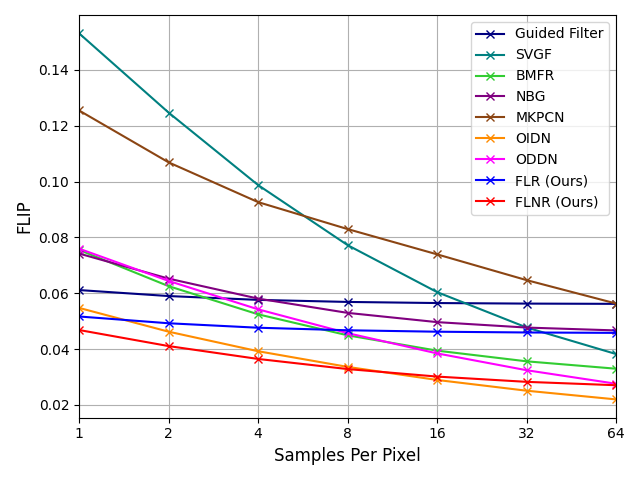}
  \end{tabular}
  \caption{Evaluation of quality metrics for increasing spp budgets.}
  \label{fig:image_quality}
\end{figure*}

\section{Evaluation}

\subsection{Dataset}

We rendered our dataset using the \texttt{Pbrt-v4} wavefront path integrator \cite{Pharr2023}.
The scenes used in the training and test sets are disjoint.
Non-Lambertian reflectance has been disabled.

\textbf{The training set} contains 300 1080p frames rendered from \texttt{The Grey \& White Room}, \texttt{The White Room}, \texttt{Contemporary Bathroom}, \texttt{Country Kitchen}, \texttt{Victorian Style House} \cite{Bitterli2016} and \texttt{Villa} \cite{Boyer2020}.
We use a 80/20 split between training and validation.

\textbf{The test set} contains 180 1080p frames rendered at the same resolution from \texttt{Sponza Remastered} \cite{Meinl2022}, \texttt{San Miguel} \cite{Llaguno2020}, \texttt{Amazon Lumberyard Bistro} \cite{Lumberyard2017} and \texttt{Japanese Classroom} \cite{Bitterli2016}. 
This is the set used in this evaluation, and for figures in this paper.

Each frame comprises:
\begin{itemize}
  \item Path traced indirect lighting at multiple sample counts (powers of 2 from 1spp to 64spp).
  \item A ground truth indirect lighting render at 4096spp.
  \item Frame metatadata comprising albedo, surface normal, depth, ambient occlusion and world-space coordinate images. These are sampled from the primary ray intersection.
  \item We also output the first (direct lighting) bounce, although it is not used in this evaluation.
\end{itemize}

The training set was only used to train our model. We use the provided weights for other neural network based methods in our evaluation. For each sample, we trace one direct and one indirect bounce. At each intersection we also trace one ray to check for light visibility, resulting in 4 rays traced in total per sample.

Ambient occlusion (AO) is rendered using \texttt{Pbrt-v4}'s ambient occlusion integrator for 64spp with distance set to 2. In a deployed solution this would instead be rasterised, which can for example be achieved by baking offline ray cast ambient occlusion into textures.

\texttt{Pbrt-v4} is an end-to-end spectral renderer: by default, each ray carries radiance for 4 discrete, randomly-sampled wavelengths which are used to estimate the color of each pixel.
For low-spp renders this additional source of randomness introduces significant additional chromatic noise into the input images.
We therefore disable wavelength jittering and increase the number of discrete wavelengths carried by each ray from 4 to 32, so that a deterministic spectral response can be approximated with just one sample.

\subsection{Compared Methods}

To evaluate our proposed methods, we compared FLR and FLNR against multiple methods designed for denoising at interactive frame rates:
\begin{itemize}
  \item The OptiX Neural Network Denoiser (ONND), which is freely available in the OptiX 7.3.0 SDK, based on the work presented by Chaitanaya et al. \cite{Chaitanya2017}.
  \item The Open Image Denoise Network (OIDN), which is freely available in the Intel Open Image Denoise Library \cite{Attila2024}.
  \item The Neural Bilateral Grid (NBG) proposed by Meng et al. \cite{Xiaoxu2020}.
  \item The Multi-Resolution Kernel Prediction CNN (MKPCN) \cite{Xiaoxu2020}, which is a multi-resolution variant of the Kernel prediction network proposed by Bako et al. \cite{Bako2017} optimised for real-time inference.
  \item The guided filter implementation proposed by Bauszat for global illumination denoising \cite{Bauszat2011}.
  \item The Blockwise Multi-Order Feature Regression (BMFR) denoiser \cite{Koskela2019}.
  \item The Spatiotemporal Variance Guided Filtering (SVGF) \cite{Schied2017}.
\end{itemize}

For ONND and OIDN, we used the denoiser implementations provided in the respective libraries.
For NBG, MKPCN and BMFR, we used the original code implementations provided by the authors with the paper. 
For NBG and MKPCN we also use the provided weights pre-trained on the BMFR dataset \cite{Xiaoxu2020}.
The guided filter was implemented in Pytorch following the original paper by He et al.~\cite{He2010} and using a radius of 24 and an epsilon of 0.01 as specified in Bauszat's paper~\cite{Bauszat2011}. 
As the authors of SVGF did not provide a code implementation of their method, we followed the BMFR authors and used the code implementation they used for their evaluation (https://github.com/ruba/RadeonProRender-Baikal).

\subsection{Quantitative Results}

We compare the quality of single-frame indirect lighting reconstruction and execution time (Nvidia RTX 2080 Ti) of several methods in Table~\ref{tab:results}. While we have tried to make this comparison as fair as possible, we note that:
\begin{itemize}
\item OIDN~\cite{Attila2024} and ODDN~\cite{Chaitanya2017} are neural network denoisers designed for application to non-Lambertian scenes.
\item BMFR~\cite{Koskela2019} and SVGF~\cite{Schied2017} are spatiotemporal solutions expecting a sequence of frames, and would achieve better results under those conditions.
\end{itemize}
For consistency, we treat all denoisers as functions of the form:

\begin{equation}
I = f(P, G, A)
\end{equation}

Where $I$ is the denoised indirect lighting output modulated with albedo, $P$ is the noisy indirect lighting modulated with albedo, $G$ is the set of auxiliary inputs or guides (please refer to Table~\ref{tab:results} for each method), and $A$ is the albedo. Where the method is designed to denoise lighting separately from albedo (as in our case), we divide $P$ by $A$, denoise, then multiply again with $A$ before outputting. OIDN and ODDN instead operate on the albedo-modulated input, with albedo as an auxiliary input.

We compare the 1spp denoised outputs against the 4096spp ground truth using a range of metrics: PSNR, SSIM~\cite{Wang2004}, rMSE and FLIP~\cite{Pontus2020} in Table~\ref{tab:results}. Our full FLNR method, including the ambient occlusion guide channel and neural network, achieves highest result quality across all metrics. We also include results for FLR, with and without ambient occlusion. Image quality as a function of input sampling rate is shown in Figure~\ref{fig:image_quality}, showing that other methods do achieve higher quality on this task at higher sampling rates.

\begin{figure}[t]
  \centering
  \includegraphics[width=0.9\columnwidth]{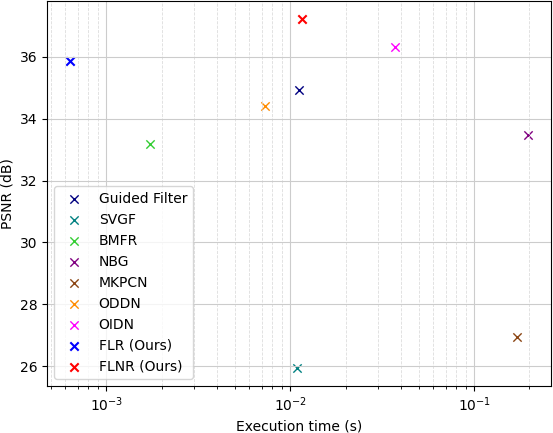}
  \caption{
    Denoising quality (PSNR) for each method when denoising 1spp input images, against execution time. Our proposed methods achieve both good quality and fast execution. Fast, high-quality methods appear towards the top-left of the plot. Note the logarithmic x-axis.
  }
  \label{fig:quality_perf}
\end{figure}

As well as maximising result quality, we also achieve faster execution times than comparable methods, as shown in Figure~\ref{fig:quality_perf}. Our proposed methods target two interesting performance points: high quality at fastest execution (FLR), and maximising quality whilst remaining faster-than-real-time (FLNR).

\begin{figure*}[h!]
  \centering
  \includegraphics[width=1\linewidth]{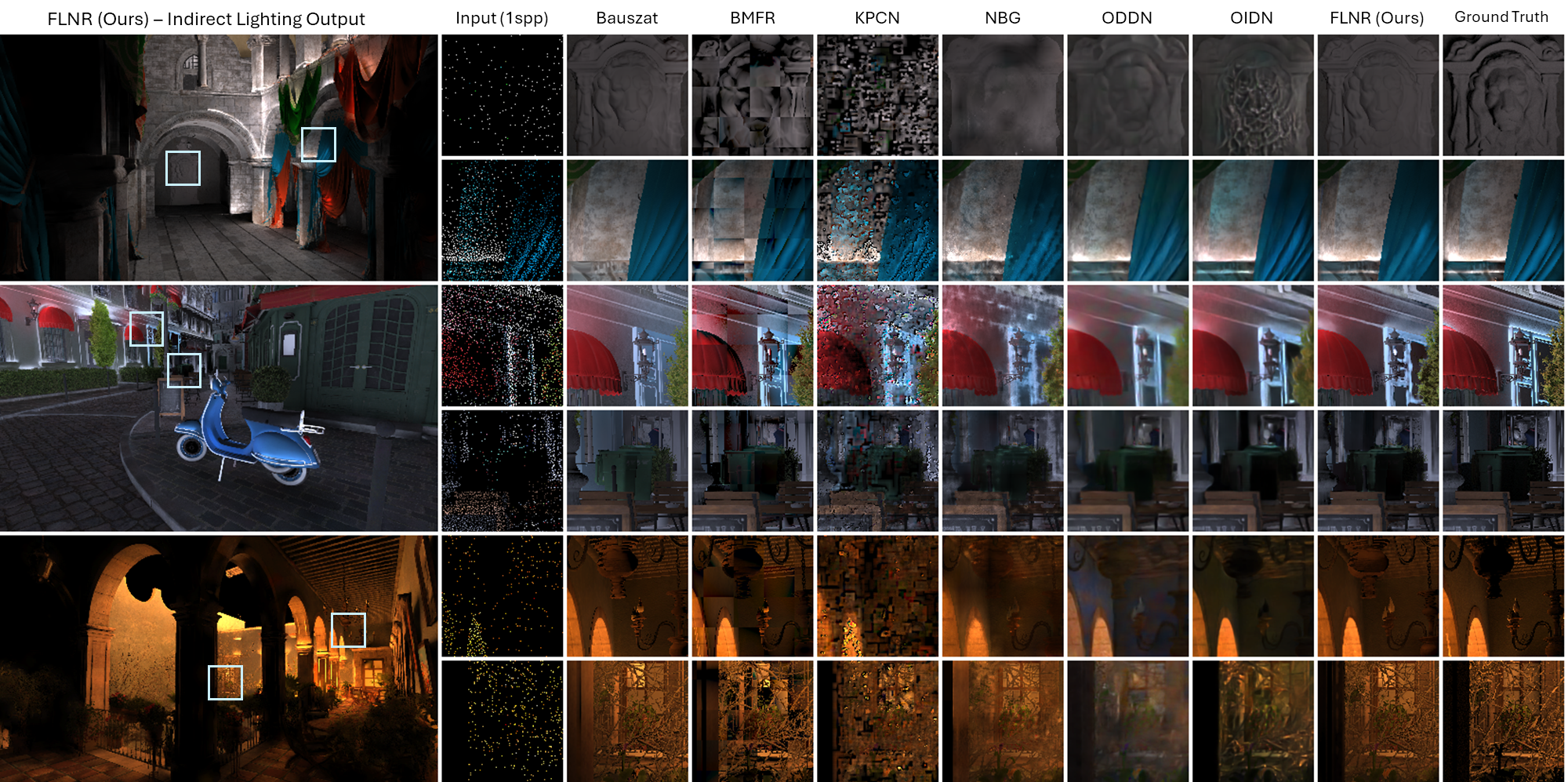}
  \caption{
    Visual comparison of albedo-modulated results for denoising of 1spp indirect lighting. Locations of crops are indicated in the full frames on the left.
  }
  \label{fig:thumbnails}
\end{figure*}

\subsection{Qualitative Results}

Figure~\ref{fig:thumbnails} contains a side-by-side comparison of visual results. As a local linear method, the guided filter~\cite{Bauszat2011} generates clean, smooth outputs, but has a tendency to bleed across edges and miss cast shadows. Conversely, deep-learning based methods (OIDN~\cite{Attila2024} and ODDN~\cite{Chaitanya2017}) reconstruct cast shadows accurately, but suffer from chromatic noise and tend to overblur outputs. FLNR combines the strengths of both techniques, minimising bleeding and achieving high-quality shadow reconsruction.

\begin{figure}[t]
  \centering
  \includegraphics[width=1\linewidth]{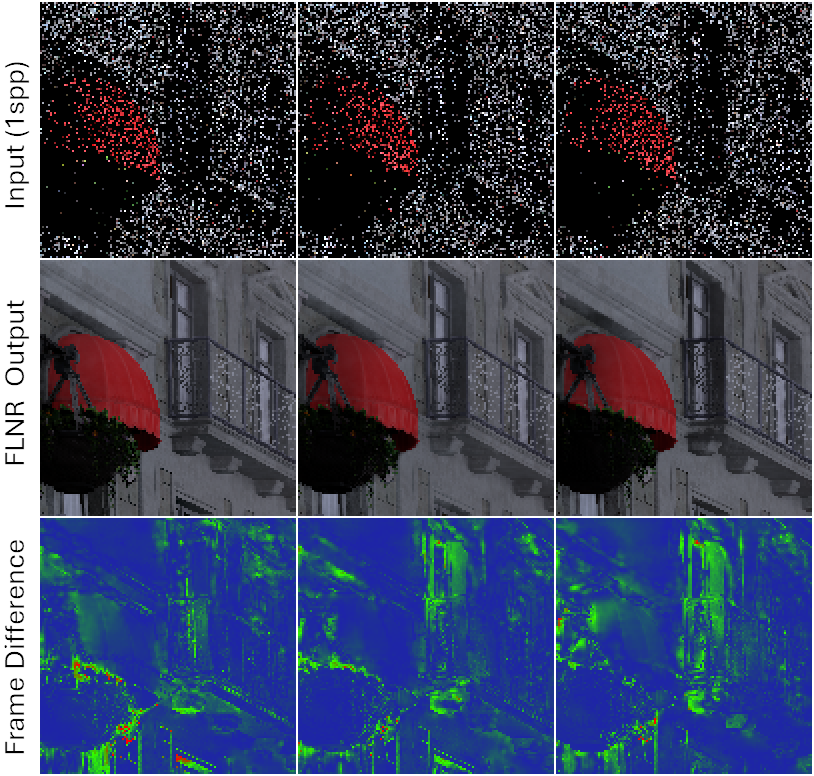}
  \caption{
    Although plausible in isolation, a sequence of indepenently denoised frames with different lighting inputs will have differing lighting reconstructions. This manifests as temporal jitter.
  }
  \label{fig:jitter}
\end{figure}

\section{Conclusions and Future Work}

We presented a novel means of estimating converged light transport of path-traced Lambertian scenes from very low sample counts.
We identified that local linear models, with their excellent computational efficiency, sharp output and applicability to high-noise inputs, naturally complement the strengths of neural networks.
Using a neural network to extract structural information from the input guide channels yielded outputs with higher visual quality than is achievable with either a local linear model or a comparably-sized neural network in isolation.
Our solution builds on this insight, allowing us to maximise result quality at constrained ray counts typical of desktop and mobile GPUs, making it suitable for implementation as part of a real-time rendering solution on commodity GPU hardware.

As a pure spatial denoiser, FLNR is subject to jitter artefacts when applied to temporal sequences (Figure~\ref{fig:jitter}).
We expect temporal jitter to be addressed effectively by inclusion of temporal filtering, which will also result in a further improvement in result quality.
Robust single-frame denoising, such as we have demonstrated, is an excellent foundation for building such a spatiotemporal filter, since it affords greater reactivity and more rapid convergence following changes in viewpoint and lighting.

Although many scenes are Lambertian to a first approximation, this restriction of our technique should be relaxed in future work. We anticipate that it can form the core of a more comprehensive denoiser capable of handling more complex surface reflectance behaviour.

\section*{Acknowledgements}

The authors would like to thank David Walton, Cagatay Dikici, Insu Yu, Alan Wolfe and Martin Roberts for their input at various stages of this project's development. 
We are also grateful to Mathieu Einig for his considerable help with prototyping and testing.

\bibliographystyle{ACM-Reference-Format}
\bibliography{denoiser_bibliography}


\begin{thebibliography}{39}


\ifx \showCODEN    \undefined \def \showCODEN     #1{\unskip}     \fi
\ifx \showDOI      \undefined \def \showDOI       #1{#1}\fi
\ifx \showISBNx    \undefined \def \showISBNx     #1{\unskip}     \fi
\ifx \showISBNxiii \undefined \def \showISBNxiii  #1{\unskip}     \fi
\ifx \showISSN     \undefined \def \showISSN      #1{\unskip}     \fi
\ifx \showLCCN     \undefined \def \showLCCN      #1{\unskip}     \fi
\ifx \shownote     \undefined \def \shownote      #1{#1}          \fi
\ifx \showarticletitle \undefined \def \showarticletitle #1{#1}   \fi
\ifx \showURL      \undefined \def \showURL       {\relax}        \fi
\providecommand\bibfield[2]{#2}
\providecommand\bibinfo[2]{#2}
\providecommand\natexlab[1]{#1}
\providecommand\showeprint[2][]{arXiv:#2}

\bibitem[\protect\citeauthoryear{{\'A}fra}{{\'A}fra}{2024}]%
        {Attila2024}
\bibfield{author}{\bibinfo{person}{Attila~T. {\'A}fra}.}
  \bibinfo{year}{2024}\natexlab{}.
\newblock \bibinfo{title}{{Intel\textsuperscript{\textregistered} Open Image
  Denoise}}.
\newblock
\newblock
\newblock
\shownote{\url{https://www.openimagedenoise.org}.}


\bibitem[\protect\citeauthoryear{Andersson, Nilsson, Akenine-M\"{o}ller,
  Oskarsson, \r{A}str\"{o}m, and Fairchild}{Andersson et~al\mbox{.}}{2020}]%
        {Pontus2020}
\bibfield{author}{\bibinfo{person}{Pontus Andersson}, \bibinfo{person}{Jim
  Nilsson}, \bibinfo{person}{Tomas Akenine-M\"{o}ller}, \bibinfo{person}{Magnus
  Oskarsson}, \bibinfo{person}{Kalle \r{A}str\"{o}m}, {and}
  \bibinfo{person}{Mark~D. Fairchild}.} \bibinfo{year}{2020}\natexlab{}.
\newblock \showarticletitle{FLIP: A Difference Evaluator for Alternating
  Images}.
\newblock \bibinfo{journal}{\emph{Proc. ACM Comput. Graph. Interact. Tech.}}
  \bibinfo{volume}{3}, \bibinfo{number}{2}, Article \bibinfo{articleno}{15}
  (\bibinfo{date}{aug} \bibinfo{year}{2020}), \bibinfo{numpages}{23}~pages.
\newblock
\urldef\tempurl%
\url{https://doi.org/10.1145/3406183}
\showDOI{\tempurl}


\bibitem[\protect\citeauthoryear{Bako, Vogels, Mcwilliams, Meyer, Nov\'{a}K,
  Harvill, Sen, Derose, and Rousselle}{Bako et~al\mbox{.}}{2017}]%
        {Bako2017}
\bibfield{author}{\bibinfo{person}{Steve Bako}, \bibinfo{person}{Thijs Vogels},
  \bibinfo{person}{Brian Mcwilliams}, \bibinfo{person}{Mark Meyer},
  \bibinfo{person}{Jan Nov\'{a}K}, \bibinfo{person}{Alex Harvill},
  \bibinfo{person}{Pradeep Sen}, \bibinfo{person}{Tony Derose}, {and}
  \bibinfo{person}{Fabrice Rousselle}.} \bibinfo{year}{2017}\natexlab{}.
\newblock \showarticletitle{Kernel-predicting convolutional networks for
  denoising Monte Carlo renderings}.
\newblock \bibinfo{journal}{\emph{ACM Trans. Graph.}} \bibinfo{volume}{36},
  \bibinfo{number}{4}, Article \bibinfo{articleno}{97} (\bibinfo{date}{jul}
  \bibinfo{year}{2017}), \bibinfo{numpages}{14}~pages.
\newblock
\showISSN{0730-0301}
\urldef\tempurl%
\url{https://doi.org/10.1145/3072959.3073708}
\showDOI{\tempurl}


\bibitem[\protect\citeauthoryear{Balint, Wolski, Myszkowski, Seidel, and
  Mantiuk}{Balint et~al\mbox{.}}{2023}]%
        {Balint2023}
\bibfield{author}{\bibinfo{person}{Martin Balint}, \bibinfo{person}{Krzysztof
  Wolski}, \bibinfo{person}{Karol Myszkowski}, \bibinfo{person}{Hans-Peter
  Seidel}, {and} \bibinfo{person}{Rafa\l{} Mantiuk}.}
  \bibinfo{year}{2023}\natexlab{}.
\newblock \showarticletitle{Neural Partitioning Pyramids for Denoising Monte
  Carlo Renderings}. In \bibinfo{booktitle}{\emph{ACM SIGGRAPH 2023 Conference
  Proceedings}} (, Los Angeles, CA, USA,) \emph{(\bibinfo{series}{SIGGRAPH
  '23})}. \bibinfo{publisher}{Association for Computing Machinery},
  \bibinfo{address}{New York, NY, USA}, Article \bibinfo{articleno}{60},
  \bibinfo{numpages}{11}~pages.
\newblock
\showISBNx{9798400701597}
\urldef\tempurl%
\url{https://doi.org/10.1145/3588432.3591562}
\showDOI{\tempurl}


\bibitem[\protect\citeauthoryear{Bauszat, Eisemann, and Magnor}{Bauszat
  et~al\mbox{.}}{2011}]%
        {Bauszat2011}
\bibfield{author}{\bibinfo{person}{Pablo Bauszat}, \bibinfo{person}{Martin
  Eisemann}, {and} \bibinfo{person}{Marcus Magnor}.}
  \bibinfo{year}{2011}\natexlab{}.
\newblock \showarticletitle{Guided Image Filtering for Interactive High-quality
  Global Illumination}.
\newblock \bibinfo{journal}{\emph{Computer Graphics Forum}}
  \bibinfo{volume}{30}, \bibinfo{number}{4} (\bibinfo{year}{2011}),
  \bibinfo{pages}{1361--1368}.
\newblock
\urldef\tempurl%
\url{https://doi.org/10.1111/j.1467-8659.2011.01996.x}
\showDOI{\tempurl}
\showeprint{https://onlinelibrary.wiley.com/doi/pdf/10.1111/j.1467-8659.2011.01996.x}


\bibitem[\protect\citeauthoryear{Bitterli}{Bitterli}{2016}]%
        {Bitterli2016}
\bibfield{author}{\bibinfo{person}{Benedikt Bitterli}.}
  \bibinfo{year}{2016}\natexlab{}.
\newblock \bibinfo{title}{Rendering resources}.
\newblock
\newblock
\newblock
\shownote{https://benedikt-bitterli.me/resources/.}


\bibitem[\protect\citeauthoryear{Boyer}{Boyer}{2020}]%
        {Boyer2020}
\bibfield{author}{\bibinfo{person}{Florent Boyer}.}
  \bibinfo{year}{2020}\natexlab{}.
\newblock \bibinfo{title}{pbrt-v4-scenes: Villa}.
\newblock
\newblock
\newblock
\shownote{https://github.com/mmp/pbrt-v4-scenes.}


\bibitem[\protect\citeauthoryear{Briedis, Djelouah, Ortiz, Meyer, Gross, and
  Schroers}{Briedis et~al\mbox{.}}{2023}]%
        {Briedis2023}
\bibfield{author}{\bibinfo{person}{Karlis~Martins Briedis},
  \bibinfo{person}{Abdelaziz Djelouah}, \bibinfo{person}{Rapha\"{e}l Ortiz},
  \bibinfo{person}{Mark Meyer}, \bibinfo{person}{Markus Gross}, {and}
  \bibinfo{person}{Christopher Schroers}.} \bibinfo{year}{2023}\natexlab{}.
\newblock \showarticletitle{Kernel-Based Frame Interpolation for
  Spatio-Temporally Adaptive Rendering}. In \bibinfo{booktitle}{\emph{ACM
  SIGGRAPH 2023 Conference Proceedings}} (, Los Angeles, CA, USA,)
  \emph{(\bibinfo{series}{SIGGRAPH '23})}. \bibinfo{publisher}{Association for
  Computing Machinery}, \bibinfo{address}{New York, NY, USA}, Article
  \bibinfo{articleno}{59}, \bibinfo{numpages}{11}~pages.
\newblock
\showISBNx{9798400701597}
\urldef\tempurl%
\url{https://doi.org/10.1145/3588432.3591497}
\showDOI{\tempurl}


\bibitem[\protect\citeauthoryear{Chaitanya, Kaplanyan, Schied, Salvi, Lefohn,
  Nowrouzezahrai, and Aila}{Chaitanya et~al\mbox{.}}{2017}]%
        {Chaitanya2017}
\bibfield{author}{\bibinfo{person}{Chakravarty R.~Alla Chaitanya},
  \bibinfo{person}{Anton~S. Kaplanyan}, \bibinfo{person}{Christoph Schied},
  \bibinfo{person}{Marco Salvi}, \bibinfo{person}{Aaron Lefohn},
  \bibinfo{person}{Derek Nowrouzezahrai}, {and} \bibinfo{person}{Timo Aila}.}
  \bibinfo{year}{2017}\natexlab{}.
\newblock \showarticletitle{Interactive reconstruction of Monte Carlo image
  sequences using a recurrent denoising autoencoder}.
\newblock \bibinfo{journal}{\emph{ACM Trans. Graph.}} \bibinfo{volume}{36},
  \bibinfo{number}{4}, Article \bibinfo{articleno}{98} (\bibinfo{date}{jul}
  \bibinfo{year}{2017}), \bibinfo{numpages}{12}~pages.
\newblock
\showISSN{0730-0301}
\urldef\tempurl%
\url{https://doi.org/10.1145/3072959.3073601}
\showDOI{\tempurl}


\bibitem[\protect\citeauthoryear{Chowdhury, Kawiak, de~Boer, and
  Xavier}{Chowdhury et~al\mbox{.}}{2022}]%
        {Chowdhury2022}
\bibfield{author}{\bibinfo{person}{Hisham Chowdhury},
  \bibinfo{person}{Robert~Rense Kawiak}, \bibinfo{person}{Gabriel~Ferreira de
  Boer}, {and} \bibinfo{person}{Lucas Xavier}.}
  \bibinfo{year}{2022}\natexlab{}.
\newblock \bibinfo{title}{Intel XeSS – an AI based Super Sampling solution
  for Real-time Rendering}.  (\bibinfo{year}{2022}).
\newblock
\urldef\tempurl%
\url{https://www.youtube.com/watch?v=fkOaT6SWX0w}
\showURL{%
\tempurl}
\newblock
\shownote{In Game Developers Conference.}


\bibitem[\protect\citeauthoryear{Cleveland}{Cleveland}{1979}]%
        {cleveland1979robust}
\bibfield{author}{\bibinfo{person}{William~S Cleveland}.}
  \bibinfo{year}{1979}\natexlab{}.
\newblock \showarticletitle{Robust locally weighted regression and smoothing
  scatterplots}.
\newblock \bibinfo{journal}{\emph{Journal of the American statistical
  association}} \bibinfo{volume}{74}, \bibinfo{number}{368}
  (\bibinfo{year}{1979}), \bibinfo{pages}{829--836}.
\newblock


\bibitem[\protect\citeauthoryear{Dammertz, Sewtz, Hanika, and Lensch}{Dammertz
  et~al\mbox{.}}{2010}]%
        {Dammertz2010}
\bibfield{author}{\bibinfo{person}{Holger Dammertz}, \bibinfo{person}{Daniel
  Sewtz}, \bibinfo{person}{Johannes Hanika}, {and} \bibinfo{person}{Hendrik
  P.~A. Lensch}.} \bibinfo{year}{2010}\natexlab{}.
\newblock \showarticletitle{Edge-avoiding \`{A}-Trous wavelet transform for
  fast global illumination filtering}. In \bibinfo{booktitle}{\emph{Proceedings
  of the Conference on High Performance Graphics}} (Saarbrucken, Germany)
  \emph{(\bibinfo{series}{HPG '10})}. \bibinfo{publisher}{Eurographics
  Association}, \bibinfo{address}{Goslar, DEU}, \bibinfo{pages}{67–75}.
\newblock


\bibitem[\protect\citeauthoryear{Hasselgren, Munkberg, Salvi, Patney, and
  Lefohn}{Hasselgren et~al\mbox{.}}{2020}]%
        {Hasselgren2020}
\bibfield{author}{\bibinfo{person}{J. Hasselgren}, \bibinfo{person}{J.
  Munkberg}, \bibinfo{person}{M. Salvi}, \bibinfo{person}{A. Patney}, {and}
  \bibinfo{person}{A. Lefohn}.} \bibinfo{year}{2020}\natexlab{}.
\newblock \showarticletitle{Neural Temporal Adaptive Sampling and Denoising}.
\newblock \bibinfo{journal}{\emph{Computer Graphics Forum}}
  \bibinfo{volume}{39}, \bibinfo{number}{2} (\bibinfo{year}{2020}),
  \bibinfo{pages}{147--155}.
\newblock
\urldef\tempurl%
\url{https://doi.org/10.1111/cgf.13919}
\showDOI{\tempurl}
\showeprint{https://onlinelibrary.wiley.com/doi/pdf/10.1111/cgf.13919}


\bibitem[\protect\citeauthoryear{He and Sun}{He and Sun}{2015}]%
        {He2015}
\bibfield{author}{\bibinfo{person}{Kaiming He} {and} \bibinfo{person}{Jian
  Sun}.} \bibinfo{year}{2015}\natexlab{}.
\newblock \bibinfo{title}{Fast Guided Filter}.  (\bibinfo{year}{2015}).
\newblock
\newblock
\shownote{arXiv:1505.00996.}


\bibitem[\protect\citeauthoryear{He, Sun, and Tang}{He et~al\mbox{.}}{2010}]%
        {He2010}
\bibfield{author}{\bibinfo{person}{Kaiming He}, \bibinfo{person}{Jian Sun},
  {and} \bibinfo{person}{Xiaoou Tang}.} \bibinfo{year}{2010}\natexlab{}.
\newblock \showarticletitle{Guided Image Filtering}. In
  \bibinfo{booktitle}{\emph{Computer Vision -- ECCV 2010}},
  \bibfield{editor}{\bibinfo{person}{Kostas Daniilidis},
  \bibinfo{person}{Petros Maragos}, {and} \bibinfo{person}{Nikos Paragios}}
  (Eds.). \bibinfo{publisher}{Springer Berlin Heidelberg},
  \bibinfo{address}{Berlin, Heidelberg}, \bibinfo{pages}{1--14}.
\newblock
\showISBNx{978-3-642-15549-9}


\bibitem[\protect\citeauthoryear{I\c{s}\i{}k, Mullia, Fisher, Eisenmann, and
  Gharbi}{I\c{s}\i{}k et~al\mbox{.}}{2021}]%
        {Isik2021}
\bibfield{author}{\bibinfo{person}{Mustafa I\c{s}\i{}k},
  \bibinfo{person}{Krishna Mullia}, \bibinfo{person}{Matthew Fisher},
  \bibinfo{person}{Jonathan Eisenmann}, {and} \bibinfo{person}{Micha\"{e}l
  Gharbi}.} \bibinfo{year}{2021}\natexlab{}.
\newblock \showarticletitle{Interactive Monte Carlo denoising using affinity of
  neural features}.
\newblock \bibinfo{journal}{\emph{ACM Trans. Graph.}} \bibinfo{volume}{40},
  \bibinfo{number}{4}, Article \bibinfo{articleno}{37} (\bibinfo{date}{jul}
  \bibinfo{year}{2021}), \bibinfo{numpages}{13}~pages.
\newblock
\showISSN{0730-0301}
\urldef\tempurl%
\url{https://doi.org/10.1145/3450626.3459793}
\showDOI{\tempurl}


\bibitem[\protect\citeauthoryear{Kajiya}{Kajiya}{1986}]%
        {Kajiya1986}
\bibfield{author}{\bibinfo{person}{James~T. Kajiya}.}
  \bibinfo{year}{1986}\natexlab{}.
\newblock \showarticletitle{The rendering equation}.
\newblock \bibinfo{journal}{\emph{SIGGRAPH Comput. Graph.}}
  \bibinfo{volume}{20}, \bibinfo{number}{4} (\bibinfo{date}{aug}
  \bibinfo{year}{1986}), \bibinfo{pages}{143–150}.
\newblock
\showISSN{0097-8930}
\urldef\tempurl%
\url{https://doi.org/10.1145/15886.15902}
\showDOI{\tempurl}


\bibitem[\protect\citeauthoryear{Kalantari, Bako, and Sen}{Kalantari
  et~al\mbox{.}}{2015}]%
        {Kalantari2015}
\bibfield{author}{\bibinfo{person}{Nima~Khademi Kalantari},
  \bibinfo{person}{Steve Bako}, {and} \bibinfo{person}{Pradeep Sen}.}
  \bibinfo{year}{2015}\natexlab{}.
\newblock \showarticletitle{A machine learning approach for filtering Monte
  Carlo noise}.
\newblock \bibinfo{journal}{\emph{ACM Trans. Graph.}} \bibinfo{volume}{34},
  \bibinfo{number}{4}, Article \bibinfo{articleno}{122} (\bibinfo{date}{jul}
  \bibinfo{year}{2015}), \bibinfo{numpages}{12}~pages.
\newblock
\showISSN{0730-0301}
\urldef\tempurl%
\url{https://doi.org/10.1145/2766977}
\showDOI{\tempurl}


\bibitem[\protect\citeauthoryear{Koskela, Immonen, M\"{a}kitalo, Foi, Viitanen,
  J\"{a}\"{a}skel\"{a}inen, Kultala, and Takala}{Koskela et~al\mbox{.}}{2019}]%
        {Koskela2019}
\bibfield{author}{\bibinfo{person}{Matias Koskela}, \bibinfo{person}{Kalle
  Immonen}, \bibinfo{person}{Markku M\"{a}kitalo}, \bibinfo{person}{Alessandro
  Foi}, \bibinfo{person}{Timo Viitanen}, \bibinfo{person}{Pekka
  J\"{a}\"{a}skel\"{a}inen}, \bibinfo{person}{Heikki Kultala}, {and}
  \bibinfo{person}{Jarmo Takala}.} \bibinfo{year}{2019}\natexlab{}.
\newblock \showarticletitle{Blockwise Multi-Order Feature Regression for
  Real-Time Path-Tracing Reconstruction}.
\newblock \bibinfo{journal}{\emph{ACM Trans. Graph.}} \bibinfo{volume}{38},
  \bibinfo{number}{5}, Article \bibinfo{articleno}{138} (\bibinfo{date}{jun}
  \bibinfo{year}{2019}), \bibinfo{numpages}{14}~pages.
\newblock
\showISSN{0730-0301}
\urldef\tempurl%
\url{https://doi.org/10.1145/3269978}
\showDOI{\tempurl}


\bibitem[\protect\citeauthoryear{Liu}{Liu}{2020}]%
        {Liu2020}
\bibfield{author}{\bibinfo{person}{Edward Liu}.}
  \bibinfo{year}{2020}\natexlab{}.
\newblock \bibinfo{title}{DLSS 2.0 - Image Reconstruction for Real-Time
  Rendering with Deep learning}.  (\bibinfo{year}{2020}).
\newblock
\urldef\tempurl%
\url{https://www.gdcvault.com/play/1026697}
\showURL{%
\tempurl}
\newblock
\shownote{In Game Developers Conference.}


\bibitem[\protect\citeauthoryear{Liu, Llamas, Cañada, and Kelly}{Liu
  et~al\mbox{.}}{2019}]%
        {Liu2019}
\bibfield{author}{\bibinfo{person}{Edward Liu}, \bibinfo{person}{Ignacio
  Llamas}, \bibinfo{person}{Juan Cañada}, {and} \bibinfo{person}{Patrick
  Kelly}.} \bibinfo{year}{2019}\natexlab{}.
\newblock \bibinfo{booktitle}{\emph{Ray Tracing Gems}}.
\newblock \bibinfo{publisher}{Apress}. 289--319 pages.
\newblock
\newblock
\shownote{\url{http://raytracinggems.com}.}


\bibitem[\protect\citeauthoryear{Llaguno}{Llaguno}{2020}]%
        {Llaguno2020}
\bibfield{author}{\bibinfo{person}{Guillermo M.~Leal Llaguno}.}
  \bibinfo{year}{2020}\natexlab{}.
\newblock \bibinfo{title}{pbrt-v4-scenes: San Miguel}.
\newblock
\newblock
\newblock
\shownote{https://github.com/mmp/pbrt-v4-scenes.}


\bibitem[\protect\citeauthoryear{Lumberyard}{Lumberyard}{2017}]%
        {Lumberyard2017}
\bibfield{author}{\bibinfo{person}{Amazon Lumberyard}.}
  \bibinfo{year}{2017}\natexlab{}.
\newblock \bibinfo{title}{Amazon Lumberyard Bistro, Open Research Content
  Archive (ORCA)}.
\newblock
\newblock
\urldef\tempurl%
\url{http://developer.nvidia.com/orca/amazon-lumberyard-bistro}
\showURL{%
\tempurl}
\newblock
\shownote{\small
  \texttt{http://developer.nvidia.com/orca/amazon-lumberyard-bistro}.}


\bibitem[\protect\citeauthoryear{Meinl, Putica, Siqueria, Heath, Prazen,
  Herholz, Cherniak, and Kaplanyan}{Meinl et~al\mbox{.}}{2022}]%
        {Meinl2022}
\bibfield{author}{\bibinfo{person}{Frank Meinl}, \bibinfo{person}{Katica
  Putica}, \bibinfo{person}{Cristiano Siqueria}, \bibinfo{person}{Timothy
  Heath}, \bibinfo{person}{Justin Prazen}, \bibinfo{person}{Sebastian Herholz},
  \bibinfo{person}{Bruce Cherniak}, {and} \bibinfo{person}{Anton Kaplanyan}.}
  \bibinfo{year}{2022}\natexlab{}.
\newblock \bibinfo{title}{Intel Sample Library}.
\newblock
\newblock
\newblock
\shownote{https://www.intel.com/content/www/us/en/developer/topic-technology/graphics-processing-research/samples.html.}


\bibitem[\protect\citeauthoryear{Meng, Zheng, Varshney, Singh, and
  Zwicker}{Meng et~al\mbox{.}}{2020}]%
        {Xiaoxu2020}
\bibfield{author}{\bibinfo{person}{Xiaoxu Meng}, \bibinfo{person}{Quan Zheng},
  \bibinfo{person}{Amitabh Varshney}, \bibinfo{person}{Gurprit Singh}, {and}
  \bibinfo{person}{Matthias Zwicker}.} \bibinfo{year}{2020}\natexlab{}.
\newblock \showarticletitle{{Real-time Monte Carlo Denoising with the Neural
  Bilateral Grid}}. In \bibinfo{booktitle}{\emph{Eurographics Symposium on
  Rendering - DL-only Track}}, \bibfield{editor}{\bibinfo{person}{Carsten
  Dachsbacher} {and} \bibinfo{person}{Matt Pharr}} (Eds.).
  \bibinfo{publisher}{The Eurographics Association}.
\newblock
\showISBNx{978-3-03868-117-5}
\showISSN{1727-3463}
\urldef\tempurl%
\url{https://doi.org/10.2312/sr.20201133}
\showDOI{\tempurl}


\bibitem[\protect\citeauthoryear{M\"{u}ller, Rousselle, Nov\'{a}k, and
  Keller}{M\"{u}ller et~al\mbox{.}}{2021}]%
        {muller2021}
\bibfield{author}{\bibinfo{person}{Thomas M\"{u}ller}, \bibinfo{person}{Fabrice
  Rousselle}, \bibinfo{person}{Jan Nov\'{a}k}, {and} \bibinfo{person}{Alexander
  Keller}.} \bibinfo{year}{2021}\natexlab{}.
\newblock \showarticletitle{Real-Time Neural Radiance Caching for Path
  Tracing}.
\newblock \bibinfo{journal}{\emph{ACM Transactions on Graphics (SIGGRAPH)}}
  \bibinfo{volume}{40}, \bibinfo{number}{4} (\bibinfo{date}{July}
  \bibinfo{year}{2021}), \bibinfo{pages}{36:1--36:16}.
\newblock
\urldef\tempurl%
\url{https://doi.org/10.1145/3450626.3459812}
\showDOI{\tempurl}


\bibitem[\protect\citeauthoryear{Pharr, Jakob, and Humphreys}{Pharr
  et~al\mbox{.}}{2023}]%
        {Pharr2023}
\bibfield{author}{\bibinfo{person}{Matt Pharr}, \bibinfo{person}{Wenzel Jakob},
  {and} \bibinfo{person}{Greg Humphreys}.} \bibinfo{year}{2023}\natexlab{}.
\newblock \bibinfo{booktitle}{\emph{Physically Based Rendering: From Theory To
  Implementation} (\bibinfo{edition}{4} ed.)}.
\newblock \bibinfo{publisher}{The MIT Press}.
\newblock


\bibitem[\protect\citeauthoryear{Richter, Alhaija, and Koltun}{Richter
  et~al\mbox{.}}{2021}]%
        {Richter2021}
\bibfield{author}{\bibinfo{person}{Stephan~R. Richter},
  \bibinfo{person}{Hassan~Abu Alhaija}, {and} \bibinfo{person}{Vladlen
  Koltun}.} \bibinfo{year}{2021}\natexlab{}.
\newblock \showarticletitle{Enhancing Photorealism Enhancement}.
\newblock \bibinfo{journal}{\emph{IEEE Transactions on Pattern Analysis and
  Machine Intelligence}}  \bibinfo{volume}{45} (\bibinfo{year}{2021}),
  \bibinfo{pages}{1700--1715}.
\newblock
\urldef\tempurl%
\url{https://api.semanticscholar.org/CorpusID:234357913}
\showURL{%
\tempurl}


\bibitem[\protect\citeauthoryear{Ronneberger, Fischer, and Brox}{Ronneberger
  et~al\mbox{.}}{2015}]%
        {Ronneberger2015}
\bibfield{author}{\bibinfo{person}{Olaf Ronneberger}, \bibinfo{person}{Philipp
  Fischer}, {and} \bibinfo{person}{Thomas Brox}.}
  \bibinfo{year}{2015}\natexlab{}.
\newblock \showarticletitle{U-Net: Convolutional Networks for Biomedical Image
  Segmentation}. In \bibinfo{booktitle}{\emph{Medical Image Computing and
  Computer-Assisted Intervention -- MICCAI 2015}},
  \bibfield{editor}{\bibinfo{person}{Nassir Navab}, \bibinfo{person}{Joachim
  Hornegger}, \bibinfo{person}{William~M. Wells}, {and}
  \bibinfo{person}{Alejandro~F. Frangi}} (Eds.). \bibinfo{publisher}{Springer
  International Publishing}, \bibinfo{address}{Cham},
  \bibinfo{pages}{234--241}.
\newblock
\showISBNx{978-3-319-24574-4}


\bibitem[\protect\citeauthoryear{Salmi, Cséfalvay, and Imber}{Salmi
  et~al\mbox{.}}{2023}]%
        {Salmi2023}
\bibfield{author}{\bibinfo{person}{A. Salmi}, \bibinfo{person}{Sz. Cséfalvay},
  {and} \bibinfo{person}{J. Imber}.} \bibinfo{year}{2023}\natexlab{}.
\newblock \showarticletitle{Generative Adversarial Shaders for Real-Time
  Realism Enhancement}.
\newblock \bibinfo{journal}{\emph{Computer Graphics Forum}}
  \bibinfo{volume}{42}, \bibinfo{number}{8} (\bibinfo{year}{2023}),
  \bibinfo{pages}{e14870}.
\newblock
\urldef\tempurl%
\url{https://doi.org/10.1111/cgf.14870}
\showDOI{\tempurl}
\showeprint{https://onlinelibrary.wiley.com/doi/pdf/10.1111/cgf.14870}


\bibitem[\protect\citeauthoryear{Schied, Kaplanyan, Wyman, Patney, Chaitanya,
  Burgess, Liu, Dachsbacher, Lefohn, and Salvi}{Schied et~al\mbox{.}}{2017}]%
        {Schied2017}
\bibfield{author}{\bibinfo{person}{Christoph Schied}, \bibinfo{person}{Anton
  Kaplanyan}, \bibinfo{person}{Chris Wyman}, \bibinfo{person}{Anjul Patney},
  \bibinfo{person}{Chakravarty R.~Alla Chaitanya}, \bibinfo{person}{John
  Burgess}, \bibinfo{person}{Shiqiu Liu}, \bibinfo{person}{Carsten
  Dachsbacher}, \bibinfo{person}{Aaron Lefohn}, {and} \bibinfo{person}{Marco
  Salvi}.} \bibinfo{year}{2017}\natexlab{}.
\newblock \showarticletitle{Spatiotemporal variance-guided filtering: real-time
  reconstruction for path-traced global illumination}. In
  \bibinfo{booktitle}{\emph{Proceedings of High Performance Graphics}} (Los
  Angeles, California) \emph{(\bibinfo{series}{HPG '17})}.
  \bibinfo{publisher}{Association for Computing Machinery},
  \bibinfo{address}{New York, NY, USA}, Article \bibinfo{articleno}{2},
  \bibinfo{numpages}{12}~pages.
\newblock
\showISBNx{9781450351010}
\urldef\tempurl%
\url{https://doi.org/10.1145/3105762.3105770}
\showDOI{\tempurl}


\bibitem[\protect\citeauthoryear{Schied, Peters, and Dachsbacher}{Schied
  et~al\mbox{.}}{2018}]%
        {Schied2018}
\bibfield{author}{\bibinfo{person}{Christoph Schied},
  \bibinfo{person}{Christoph Peters}, {and} \bibinfo{person}{Carsten
  Dachsbacher}.} \bibinfo{year}{2018}\natexlab{}.
\newblock \showarticletitle{Gradient Estimation for Real-time Adaptive Temporal
  Filtering}.
\newblock \bibinfo{journal}{\emph{Proc. ACM Comput. Graph. Interact. Tech.}}
  \bibinfo{volume}{1}, \bibinfo{number}{2}, Article \bibinfo{articleno}{24}
  (\bibinfo{date}{aug} \bibinfo{year}{2018}), \bibinfo{numpages}{16}~pages.
\newblock
\urldef\tempurl%
\url{https://doi.org/10.1145/3233301}
\showDOI{\tempurl}


\bibitem[\protect\citeauthoryear{Thomas, Liktor, Peters, Kim, Vaidyanathan, and
  Forbes}{Thomas et~al\mbox{.}}{2022}]%
        {Thomas2022}
\bibfield{author}{\bibinfo{person}{Manu~Mathew Thomas}, \bibinfo{person}{Gabor
  Liktor}, \bibinfo{person}{Christoph Peters}, \bibinfo{person}{Sungye Kim},
  \bibinfo{person}{Karthik Vaidyanathan}, {and} \bibinfo{person}{Angus~G.
  Forbes}.} \bibinfo{year}{2022}\natexlab{}.
\newblock \showarticletitle{{Temporally Stable Real-Time Joint Neural Denoising
  and Supersampling}}. In \bibinfo{booktitle}{\emph{Proceedings of the ACM on
  Computer Graphics and Interactive Techniques}},
  \bibfield{editor}{\bibinfo{person}{Josef Spjut}, \bibinfo{person}{Marc
  Stamminger}, {and} \bibinfo{person}{Victor Zordan}} (Eds.).
  \bibinfo{publisher}{ACM Association for Computing Machinery}.
\newblock
\showISSN{2577-6193}
\urldef\tempurl%
\url{https://doi.org/10.1145/3543870}
\showDOI{\tempurl}


\bibitem[\protect\citeauthoryear{Veach}{Veach}{1998}]%
        {Veach1998}
\bibfield{author}{\bibinfo{person}{Eric Veach}.}
  \bibinfo{year}{1998}\natexlab{}.
\newblock \emph{\bibinfo{title}{Robust monte carlo methods for light transport
  simulation}}.
\newblock \bibinfo{thesistype}{Ph.D. Dissertation}. \bibinfo{address}{Stanford,
  CA, USA}.
\newblock Advisor(s) Guibas, Leonidas J.
\newblock
\showISBNx{0591907801}
\newblock
\shownote{AAI9837162.}


\bibitem[\protect\citeauthoryear{Wang, Bovik, Sheikh, and Simoncelli}{Wang
  et~al\mbox{.}}{2004}]%
        {Wang2004}
\bibfield{author}{\bibinfo{person}{Zhou Wang}, \bibinfo{person}{A.C. Bovik},
  \bibinfo{person}{H.R. Sheikh}, {and} \bibinfo{person}{E.P. Simoncelli}.}
  \bibinfo{year}{2004}\natexlab{}.
\newblock \showarticletitle{Image quality assessment: from error visibility to
  structural similarity}.
\newblock \bibinfo{journal}{\emph{IEEE Transactions on Image Processing}}
  \bibinfo{volume}{13}, \bibinfo{number}{4} (\bibinfo{year}{2004}),
  \bibinfo{pages}{600--612}.
\newblock
\urldef\tempurl%
\url{https://doi.org/10.1109/TIP.2003.819861}
\showDOI{\tempurl}


\bibitem[\protect\citeauthoryear{Zeltner, Rousselle, Weidlich, Clarberg,
  Nov\'{a}k, Bitterli, Evans, Davidovi\v{c}, Kallweit, and Lefohn}{Zeltner
  et~al\mbox{.}}{2024}]%
        {Zeltner2024}
\bibfield{author}{\bibinfo{person}{Tizian Zeltner}, \bibinfo{person}{Fabrice
  Rousselle}, \bibinfo{person}{Andrea Weidlich}, \bibinfo{person}{Petrik
  Clarberg}, \bibinfo{person}{Jan Nov\'{a}k}, \bibinfo{person}{Benedikt
  Bitterli}, \bibinfo{person}{Alex Evans}, \bibinfo{person}{Tom\'{a}\v{s}
  Davidovi\v{c}}, \bibinfo{person}{Simon Kallweit}, {and}
  \bibinfo{person}{Aaron Lefohn}.} \bibinfo{year}{2024}\natexlab{}.
\newblock \showarticletitle{Real-Time Neural Appearance Models}.
\newblock \bibinfo{journal}{\emph{ACM Trans. Graph.}} (\bibinfo{date}{apr}
  \bibinfo{year}{2024}).
\newblock
\showISSN{0730-0301}
\urldef\tempurl%
\url{https://doi.org/10.1145/3659577}
\showDOI{\tempurl}
\newblock
\shownote{Just Accepted.}


\bibitem[\protect\citeauthoryear{Zeng, Liu, Yang, Wang, and Yan}{Zeng
  et~al\mbox{.}}{2021}]%
        {Zeng2021}
\bibfield{author}{\bibinfo{person}{Zheng Zeng}, \bibinfo{person}{Shiqiu Liu},
  \bibinfo{person}{Jinglei Yang}, \bibinfo{person}{Lu Wang}, {and}
  \bibinfo{person}{Ling-Qi Yan}.} \bibinfo{year}{2021}\natexlab{}.
\newblock \showarticletitle{Temporally Reliable Motion Vectors for Real-time
  Ray Tracing}.
\newblock \bibinfo{journal}{\emph{Computer Graphics Forum}}
  \bibinfo{volume}{40}, \bibinfo{number}{2} (\bibinfo{year}{2021}),
  \bibinfo{pages}{79--90}.
\newblock
\urldef\tempurl%
\url{https://doi.org/10.1111/cgf.142616}
\showDOI{\tempurl}
\showeprint{https://onlinelibrary.wiley.com/doi/pdf/10.1111/cgf.142616}


\bibitem[\protect\citeauthoryear{Zhang and Yuan}{Zhang and Yuan}{2023}]%
        {Zhang2023}
\bibfield{author}{\bibinfo{person}{Boyu Zhang} {and} \bibinfo{person}{Hongliang
  Yuan}.} \bibinfo{year}{2023}\natexlab{}.
\newblock \showarticletitle{High-Quality Real-Time Rendering Using Subpixel
  Sampling Reconstruction}. In \bibinfo{booktitle}{\emph{AAAI Conference on
  Artificial Intelligence}}.
\newblock
\urldef\tempurl%
\url{https://api.semanticscholar.org/CorpusID:259251869}
\showURL{%
\tempurl}


\bibitem[\protect\citeauthoryear{Zwicker, Jarosz, Lehtinen, Moon, Ramamoorthi,
  Rousselle, Sen, Soler, and Yoon}{Zwicker et~al\mbox{.}}{2015}]%
        {Zwicker2015}
\bibfield{author}{\bibinfo{person}{M. Zwicker}, \bibinfo{person}{W. Jarosz},
  \bibinfo{person}{J. Lehtinen}, \bibinfo{person}{B. Moon}, \bibinfo{person}{R.
  Ramamoorthi}, \bibinfo{person}{F. Rousselle}, \bibinfo{person}{P. Sen},
  \bibinfo{person}{C. Soler}, {and} \bibinfo{person}{S.-E. Yoon}.}
  \bibinfo{year}{2015}\natexlab{}.
\newblock \showarticletitle{Recent Advances in Adaptive Sampling and
  Reconstruction for Monte Carlo Rendering}.
\newblock \bibinfo{journal}{\emph{Computer Graphics Forum}}
  \bibinfo{volume}{34}, \bibinfo{number}{2} (\bibinfo{year}{2015}),
  \bibinfo{pages}{667--681}.
\newblock
\urldef\tempurl%
\url{https://doi.org/10.1111/cgf.12592}
\showDOI{\tempurl}
\showeprint{https://onlinelibrary.wiley.com/doi/pdf/10.1111/cgf.12592}


\end{thebibliography}

\section*{Appendix}

Numerical instability and rank deficiency can easily occur due to similarity between guide
channels. This is particularly common where there are large, flat regions in the scene, and
typically results in extreme values (including \textit{NaN}s) and unacceptable artefacts.
Regularisation is therefore required when evaluating Equation~\ref{equ:closed_form}.
In this appendix, we describe our regularisation solution in detail.

To compute the matrix inverse we use the block matrix inverse
formula applied recursively:

\vspace{2pt}

\begin{footnotesize}
$ {\begin{bmatrix}\mathbf {A} &\mathbf {B} \\\mathbf {C} &\mathbf {D} \end{bmatrix}}^{-1}
={\begin{bmatrix}\mathbf {A} ^{-1}+\mathbf {A} ^{-1}\mathbf {B} \left(\mathbf {D} -\mathbf {CA} ^{-1}\mathbf {B} \right)^{-1}\mathbf {CA} ^{-1}&-\mathbf {A} ^{-1}\mathbf {B} \left(\mathbf {D} -\mathbf {CA} ^{-1}\mathbf {B} \right)^{-1}\\-\left(\mathbf {D} -\mathbf {CA} ^{-1}\mathbf {B} \right)^{-1}\mathbf {CA} ^{-1}&\left(\mathbf {D} -\mathbf {CA} ^{-1}\mathbf {B} \right)^{-1}\end{bmatrix}}$
\end{footnotesize}

\vspace{6pt}

While simple Tikhonov regularisation (Equation~\ref{equ:tikhonov}, Figure~\ref{fig:pytorch_code}) does work,
\(\varepsilon\) effectively interpolates between linear regression and a Gaussian blur, resulting in undesirable bleeding across edges.
This problem is exacerbated by the fact that each guide channel requires differing amounts of regularisation
at different points in the image.

\begin{equation}
  \textbf{A}^{*}_{\textit{k}} = \left(\textbf{X}^\intercal_{\textit{k}}\textbf{X}_{\textit{k}} + \varepsilon \textbf{I} \right)^{-1} 
  \textbf{X}^\intercal_{\textit{k}}\textbf{Y}_{\textit{k}}
\label{equ:tikhonov}
\end{equation}

This trade-off between stability and inaccuracy is the main reason why a matrix inverse is rarely used directly to solve linear
equations. More stable solvers than our block matrix inverter exist but come at the price of performance.
Therefore we designed a bespoke normalisation method to solve this issue without performance or quality trade-offs.
Our normalisation method manipulates $\textbf{X}^\intercal_{\textit{k}}\textbf{X}_{\textit{k}}$
to make Tikhonov regularisation more reliable.

Our matrix inverse would be more stable if the inputs in window
\(T_{k}\) were mean- and variance-normalised: however this is made
complicated due to the windowing function used. Luckily, since a channel of 1s among the guide channels, the (windowed)
mean and variance can be extracted from \(\textbf{X}_{k}^\intercal\textbf{X}_{k}\), and normalisation applied
directly to the moment matrices. As the first channel in \(\textbf{X}\) comprises ones, we decompose the
matrix into a scalar \(n\) in the upper left corner, a column vector
\(\textbf{u}_X^\intercal\) below it and \(\textbf{u}_X\) to its right, and a matrix \(\textbf{S}\) for the
remaining elements:

\begin{equation}
\textbf{X}_{k}^\intercal\textbf{X}_{k}=
\begin{bmatrix}
n & \begin{bmatrix}
\  & \  & \textbf{u}_X & \  & \ 
\end{bmatrix} \\
\begin{bmatrix}
\  \\
\  \\
\textbf{u}_X^\intercal \\
\  \\
\ 
\end{bmatrix} & \begin{bmatrix}
\  & \  & \  & \  & \  \\
\  & \  & \  & \  & \  \\
\  & \  & \textbf{S} & \  & \  \\
\  & \  & \  & \  & \  \\
\  & \  & \  & \  & \ 
\end{bmatrix}
\end{bmatrix}
\end{equation}

\(n\) is the (weighted) number of samples collected in the \(T_{k}\)
window if the windowing Gaussian has a maximum weight of 1, or contains
1 if the Gaussian was normalised.

$\textbf{u}_X$ contains the windowed mean $\boldsymbol{\mu}_X$ of each channel multiplied by $n$, which we can recover as:

\begin{equation}
\boldsymbol{\mu}_X = \frac{\textbf{u}_X}{n}
\end{equation}

The matrix \textbf{S} contains the windowed correlation matrix
\(\textbf{C} = \text{corr}(\textbf{X}_{k},\textbf{X}_{k})\) in the form:

\begin{equation}
\textbf{S} = \left( \boldsymbol{\sigma}_X^\intercal\boldsymbol{\sigma}_X \textbf{C} + \boldsymbol{\mu}_X^\intercal\boldsymbol{\mu}_X \right) n
\end{equation}

First, we define $\textbf{W} = \boldsymbol{\sigma}_X^\intercal\boldsymbol{\sigma}_X \textbf{C}$. Then we normalise \textbf{S} and subtract the mean:

\begin{equation}
\textbf{W} = \frac{\textbf{S}}{n} - \boldsymbol{\mu}_X^\intercal\boldsymbol{\mu}_X
\end{equation}

Then we extract the variance vector $\boldsymbol{\sigma}_X$ as:

\begin{equation}
\boldsymbol{\sigma}_X = \sqrt{\text{diag}(\textbf{W})}
\end{equation}

Now \(\textbf{C}\) can be computed as:

\begin{equation}
\textbf{C}_{ij} = \frac{\textbf{W}_{ij}}{{\boldsymbol{\sigma}_X}_{i}{\boldsymbol{\sigma}_X}_{j}}
\end{equation}

Inverting \textbf{C} is much more numerically stable than inverting $\textbf{X}^\intercal_{\textit{k}}\textbf{X}_{\textit{k}}$.
However, it is still possible to get instability if $\boldsymbol{\sigma}_X$ has a zero (or sometimes even a negative value due to numerical error).
For this reason, we reintroduce regularisation terms to the formula, with a separate additive term \(\epsilon\) and multiplicative
term \(\varepsilon\):

\begin{equation}
\widehat{\textbf{W}} = \frac{\textbf{S}}{n} + \varepsilon\ \text{diag}\left( \boldsymbol{\mu}_X^\intercal\boldsymbol{\mu}_X \right) + \epsilon\ \textbf{I}
 - (1 - \varepsilon)\boldsymbol{\mu}_X^\intercal\boldsymbol{\mu}_X
\end{equation}

From this point on we compute \textbf{C} as before:

\begin{equation}
{\widehat{\boldsymbol{\sigma}}}_{X} = \sqrt{\text{diag}\left( \widehat{\textbf{W}} \right)}
\end{equation}
\begin{equation}
{\widehat{\textbf{C}}}_{ij} = \frac{{\widehat{\textbf{W}}}_{ij}}{{{\widehat{\boldsymbol{\sigma}}}_i}{{\widehat{\boldsymbol{\sigma}}}_j}}
\end{equation}

We also need to produce a normalised version of \(\textbf{X}_{k}^\intercal\textbf{Y}_{k}\)
called \(\widehat{\textbf{B}}\). Like before, we recover
\(\boldsymbol{\mu}_{\mathbf{Y}}\) from the top row of
\(\mathbf{X}_{\text{k}}^\intercal\mathbf{Y}_{\text{k}}\) which contains
\(n\boldsymbol{\mu}_{\mathbf{Y}}\). Now we can compute
\(\widehat{\mathbf{B}}\):

\begin{equation}
{\widehat{\textbf{B}}}_{ij} = \frac{\frac{{\textbf{X}_{k}}_{i}^\intercal{\textbf{Y}_{k}}_{j}}{n} - {\boldsymbol{\mu_X}}_i{\boldsymbol{\mu_Y}_j}}{{{\widehat{\boldsymbol{\sigma}}_{X_i}}}}
\end{equation}
\begin{equation}
{\widehat{\textbf{A}}}_{k}\  = \left( {\widehat{\textbf{C}}}_{ij} + \epsilon \textbf{I} \right)^{- 1}\widehat{\textbf{B}}
\end{equation}

Now we need to substitute normalised $\textbf{x}$ values into the normalised linear
model \({\widehat{\textbf{A}}}_{k}\):

\begin{equation}
\textbf{y}_k = \frac{\textbf{x}_k - \boldsymbol{\mu}_{X}}{{\widehat{\boldsymbol{\sigma}}}_{X}}{\widehat{\textbf{A}}}_{k} + \boldsymbol{\mu}_{Y}
\end{equation}

Where \(\textbf{x}_k\) doesn't contain the first element
(which is always 1). Regularising \textbf{C} in this way gives higher quality results than directly
regularising $\textbf{X}_{k}^\intercal\textbf{X}_{k}$.
This is because it adapts our regularisation terms \(\epsilon\) and \(\varepsilon\)
to the mean and variance of the information captured by $\textbf{X}_{k}^\intercal\textbf{X}_{k}$
We learn the regularisation terms with backpropagation, which converge to values in the range
\(\left\lbrack 10^{- 5};10^{- 4} \right\rbrack\).

\end{document}